\documentclass[11pt]{article}  
\usepackage{ifpdf}

\usepackage[margin=1in]{geometry}
\usepackage[section]{placeins}
\usepackage[justification=centering]{caption}
\ifpdf
\usepackage[pdftex]{graphicx}
\usepackage{epstopdf}
\usepackage[pdftex,bookmarks=false,colorlinks=true,pdfstartview=FitBV,linkcolor=blue,citecolor=blue,urlcolor=blue]{hyperref}
\else
\usepackage{graphicx}
\usepackage{subcaption}
\usepackage{epsfig,psfrag}
\psdraft\psfull
\fi

\usepackage{amsmath} 
\usepackage{amssymb}  
\usepackage{graphicx}
\usepackage{float}
\usepackage{subfigure}
\usepackage{setspace}
\usepackage{color}
\usepackage{cite}
\usepackage{comment}
\usepackage{stackrel}
\usepackage{adjustbox}
\usepackage{tabularx}
\usepackage{tikz}
\usepackage[ruled,vlined,linesnumbered]{algorithm2e}
\usepackage{nicefrac}
\usepackage{adjustbox}

\newcommand{\dkl}{\mathrm{D}_{\mathrm{KL}}}
\newcommand{\js}{\mathrm{JS}_{\Pi}}

\newcommand{\I}{I}
\newcommand{\Q}{Q}
\renewcommand{\L}{L}
\newcommand{\q}{q}
\newcommand{\p}{p}
\renewcommand{\H}{H}
\newcommand{\tp}{{\mbox{\tiny \sf T}}}

\renewcommand{\Re}{\mathbb{R}}

\DeclareMathOperator{\argmax}{argmax}

\newtheorem{theorem}{Theorem}[section]
\newtheorem{lemma}[theorem]{Lemma}
\newtheorem{proposition}[theorem]{Proposition}

\newtheorem{definition}[theorem]{Definition}

\graphicspath{{./figures/}}

\title{\LARGE \bf
Q-Search Trees: An Information-Theoretic Approach Towards Hierarchical Abstractions for Agents with Computational Limitations
}

\author{Daniel T. Larsson%
	\thanks{D. Larsson is a PhD student with the Guggenheim School of Aerospace Engineering, Georgia Institute of Technology, Atlanta,
		GA, 30332-0150, USA. Email:
		{\small daniel.larsson@gatech.edu}}
	~~~~~~ 
	Dipankar Maity%
	\thanks{D. Maity is a Postdoctoral fellow with the Guggenheim School of Aerospace Engineering, Georgia Institute of Technology, Atlanta,
		GA, 30332-0150, USA. Email:
		{\small dmaity@gatech.edu}}
	~~~~~~ 
	Panagiotis Tsiotras%
	\thanks{P. Tsiotras is the Andrew and Lewis Chair Professor with the Guggenheim School of Aerospace Engineering and the Institute for Robotics and Intelligent Machines, Georgia Institute of Technology, Atlanta,
		GA, 30332-0150, USA. Email:
		{\small tsiotras@gatech.edu}}
}

\date{~}

\begin{document}
	
	\maketitle
	\thispagestyle{empty}
	\pagestyle{empty}

\begin{abstract}
In this paper, we develop a framework to obtain graph abstractions for decision-making by an agent where the abstractions emerge as a function of the agent's limited computational resources.
We discuss the connection of the proposed approach with information-theoretic signal compression, and formulate a novel optimization problem to obtain tree-based abstractions as a function of the agent's computational resources.
The structural properties of the new problem are discussed in detail, and two algorithmic approaches are proposed to obtain solutions to this optimization problem.
We discuss the quality of, and prove relationships between, solutions obtained by the two proposed algorithms.
The framework is demonstrated to generate a hierarchy of abstractions for a non-trivial environment.  
\end{abstract}


\section{Introduction}

Information theory provides a principled framework for obtaining optimal compressed representations of a signal \cite{Cover2006}.
The ability to form such compressed representations, also known as abstractions, has widespread uses in many fields, ranging from signal processing and data transmission, to robotic motion planning in complex environments, and many others \cite{Behnke2004,Bertsekas2012,Botvinick2012,Cover2006,Hauer2015,Hauer2016,Kambhampati1986,Kraetzschmar2004,Cowlagi2008,Cowlagi2010,Cowlagi2012,Cowlagi2012a,Einhorn2011,Tsiotras2012,Kurkoski2014,Nelson2018,Vangala2015,Larsson2017}.
Particularly for autonomous systems, simplified representations of the environment which the agent operates in are preferred, as they decrease the on-board memory requirements and reduce the computational time required to find feasible or optimal solutions for planning \cite{Cowlagi2008,Cowlagi2010,Cowlagi2012,Cowlagi2012a,Behnke2004,Hauer2015,Hauer2016,Einhorn2011,Kraetzschmar2004,Kambhampati1986,Holte2003}.

Within the realm of robotics and autonomous systems, a number of studies have leveraged the power of abstractions for both exploration and path-planning purposes.
Examples of such prior works include \cite{Cowlagi2008,Cowlagi2010,Cowlagi2012,Cowlagi2012a} in which wavelets were utilized in order to generate multi-resolution representations of two-dimensional  environments.
These compressed representations encode a simplified graph of the environment, speeding up the execution time of path-planning algorithms such as A$^*$\cite{Hauer2015}.
As the agent traverses the environment, the problem is sequentially re-solved in order to obtain a trade-off in the overall optimality of the resulting path, planning frequency, and obstacle avoidance. 
Similarly related work includes that of \cite{Hauer2015} and \cite{Hauer2016}, where the authors employed a tree-based framework in order to execute path-planning tasks in two- and three-dimensional environments.
In these studies, the planning problem involved the generation of a multi-resolution representation of the operating space of the agent in the form of a variable-depth probabilistic quadtree or octree, based on user-provided parameters and a given initial representation of the environment.
Since that framework uses probabilistic quadtrees and octrees, the initial representation of the environment is in the form of an occupancy grid, allowing for the incorporation of sensor uncertainty when creating maps of the environment \cite{Thrun2005}.

Other works have studied the generation of quadtrees in real time, such as \cite{Einhorn2011}, or the creation of multi-resolution trees from a given map and pruning rules \cite{Kraetzschmar2004}.
Abstractions have also been proposed in the reinforcement learning (RL) community in order to alleviate the curse of dimensionality, allowing for 
the solution of larger problems~\cite{Botvinick2012,Li2006}. 
However, there is no unifying method for how these abstractions are generated, as existing methods rely heavily on user-provided rules.

The drawback of all these previous works is that they do not directly address the generation of the abstractions, and instead rely on them to be either provided a-priori or created in a manner that is known beforehand. 
Furthermore, existing works do not consider the computational limitations of the agent.
That is, existing works do not consider in their formulation that agents with limited on-board resources may not employ the same representation, or depiction, of the environment as agents that are not resource limited.
The idea that all agents do not have equal capabilities has been recently discussed in the literature pertaining to the field of bounded rationality \cite{Tishby2010,Genewein2015,Ortega2011}.
In this point of view, the capabilities of an agent are represented by its information-processing abilities.
Thus, a resource-limited agent is not able to process all data collected by observing its surroundings, leading to the need for simplification of the space in which it operates. 
Utilizing these abstract representations precludes the agent from necessarily finding globally optimal solutions, but induces policies that require the agent to process fewer details of the environment in order to act \cite{Tishby2010,Lipm1995,Genewein2015,Larsson2017}.

A number of existing works have modeled single-stage and sequential bounded-rational decision making in stochastic domains by employing ideas from utility and information theory to construct constrained optimization problems \cite{Tishby2010,Larsson2017,Genewein2015,Ortega2011}.
The solution to these problems is a set of self-consistent equations, which are numerically solved by alternating iterations analogous to the Blahut-Arimoto algorithm in rate-distortion theory \cite{Genewein2015,Tishby2010,Tishby1999,Larsson2017}.
Interestingly, this framework allows for the emergence of bounded-rational policies for a range of agents with varying capabilities, recovering the rational solution in the limit \cite{Larsson2017,Genewein2015,Ortega2011,Tishby2010}.

In this paper, we address the issue of abstraction generation for a given environment, and formulate a novel optimization problem that leverages concepts from information theory to obtain representations of an environment that are a function of the agent's available resources.
Specifically, we consider the case where the environment is represented as a multi-resolution quadtree, and begin by discussing connections between environment abstractions in the form of quadtrees and general signal compression, the latter of which has been extensively studied by information theorists.
We then formulate an optimization problem over the space of trees that utilizes concepts from the information bottleneck method~\cite{Tishby1999}, 
and we subsequently propose two algorithms to solve the problem.
Theoretical guarantees of our proposed algorithmic approaches are presented and discussed.
The approach is applied to a non-trivial example, where we examine the results and discuss the interpretation of the theory as applied to bounded-rational agents.

The remainder of the paper is organized as follows.
In Section~\ref{sec:Prelims} we introduce and review the fundamental concepts needed in this work as well as  we review the connections between quadtrees and optimal signal compression.
Then, in Section~\ref{sec:Prob_formulation}, we formulate our problem and show how principles from information theory can be incorporated into a new optimization problem over the space of trees.
In Section \ref{sec:algorithms}, we propose two algorithms that can be used to solve the optimization problem and present the theoretical contributions of the paper.
Section~\ref{sec:Results_and_Discuss} presents results of the proposed methodology applied to an occupancy grid with and without prior information.
We conclude with several remarks in Section~\ref{sec:Conclusions}.


\section{Preliminaries} \label{sec:Prelims}

\subsection{Quadtree Decompositions}

We consider the emergence of abstractions in the form of multi-resolution quadtree representations.
Quadtrees are a common tool utilized in the robotics community to reduce the complexity of environments in order to speed path-planning or ease internal storage requirements \cite{Hauer2015,Hauer2016,Behnke2004,Einhorn2011,Nelson2018}.
The theoretical contributions of the paper are applicable however for any tree structure, beyond  just quadtrees.
To this end, we assume that the environment $\mathcal{W} \subset \Re^2$ (generalizable to $\Re^d$) is given by a two-dimensional grid world where each grid element is a unit square (hypercube).
We assume that there exists an integer $\ell > 0$ such that $\mathcal{W}$ is contained within a square (hypercube) of side length $2^{\ell}$.
A tree representation $\mathcal{T} = \left(\mathcal{N}, \mathcal{E}\right) = \left(\mathcal{N}(\mathcal{T}), \mathcal{E}(\mathcal{T}) \right)$ of $\mathcal{W}$ consists of a set of nodes $\mathcal{N}$ and edges $\mathcal{E}$ describing the interconnections between the nodes in the tree \cite{Hauer2016}.
We denote the set of all possible quadtree representations of maximum depth $\ell$ of $\mathcal{W}$ by $\mathcal T^{\mathcal{Q}}$ and let $\mathcal T_{\mathcal W} \in \mathcal T^{\mathcal Q}$ denote the finest quadtree representation of $\mathcal W$; an example is shown in 
Figure~\ref{fig:worldQuadtree}.
It should be noted that $\mathcal T_{\mathcal W}$ encodes a specific structure for $\mathcal W$, which we make precise in the following definition.
\begin{definition}
	\normalfont{Let $t \in \mathcal N(\mathcal T_{\mathcal W})$ be any node at depth $k \in \left\{0,\ldots,\ell \right\}$.
	Then $t' \in \mathcal N(\mathcal T_{\mathcal W})$ is a \emph{child} of $t$ if the following hold:
	\begin{enumerate}
		\item 
		Node $t'$ is at depth $k+1$ in $\mathcal T_{\mathcal W}$.
		
		\item
		Nodes $t$ and $t'$ are incident to a common edge, i.e., $\left(t,t'\right) \in \mathcal E \left( \mathcal T_{\mathcal W} \right)$.
	\end{enumerate}
	Conversely, we say that $t$ is the \emph{parent} of $t'$ if $t'$ is a child of $t$.  
	Furthermore, we let 
	\begin{equation*}
	\mathcal N_k(\mathcal T_{q}) = \left\{t \in \mathcal N(\mathcal T_{q}) : t \text{~is at depth~} k~\text{in}~\mathcal T_{\mathcal W} \right\},
	\end{equation*}
	to be the set of all nodes of the tree $\mathcal T_{q} \in \mathcal T^{\mathcal Q}$ at depth $k$.
	}
\end{definition}
\noindent We will frequently seek to relate nodes in the tree $\mathcal T_q$ to those in the tree $\mathcal T_{\mathcal W}$, which leads us to the following definition.
\begin{definition}  \label{def:nodeProp}
	\normalfont{
	Let $t \in \mathcal N(\mathcal T_{q})$ be any node in the tree $\mathcal T_{q} \in \mathcal T^{\mathcal Q}$.  Then the following hold:
	
	\begin{enumerate}
		\item 	
		The node $t$ has children 
		\begin{equation*}
		\mathcal C(t) = \left\{t' \in \mathcal N(\mathcal T_{\mathcal W}): t' \text{~is a child of~} t  \right\}.
		\end{equation*}
		\item
		The node $t$ has parent
		\begin{equation*}
		\mathcal P(t) = \left\{\hat t \in \mathcal N(\mathcal T_{\mathcal W}): t \in \mathcal C(\hat t)\right\}.
		\end{equation*}
		
		\item 
		The node $t$ is the root of the tree $\mathcal T_{q}$, denoted by $\mathrm{Root} \left( \mathcal T_q\right)$, if $\mathcal P(t) = \emptyset$.
		
		\item 
		The node $t$ is a leaf of $\mathcal T_{q}$ if $\mathcal C(t) \cap \mathcal N(\mathcal T_{q}) = \emptyset$.  
		Furthermore, the set of leaf nodes of $\mathcal T_{q}$ is given by
		\begin{equation*}
		\mathcal N_{\text{leaf}}\left( \mathcal T_{q} \right) = \left\{t' \in \mathcal N(\mathcal T_{q}) :  \mathcal C(t') \cap \mathcal N(\mathcal T_{q}) = \emptyset \right\}.
		\end{equation*}
		
		\item 
		If $t \notin \mathcal N_{\text{leaf}}( \mathcal T_{q})$ then $t \in \mathcal N_{\text{int}}( \mathcal T_{q}) =  \mathcal N(\mathcal T_{q}) \setminus  \mathcal N_{\text{leaf}}( \mathcal T_{q})$, where $\mathcal N_{\text{int}}( \mathcal T_{q})$ is the set of interior nodes of $\mathcal T_{q}$.
	\end{enumerate}
	}
\end{definition}
\noindent Note that the space $\mathcal T^{\mathcal Q}$ encodes a specific structure on the abstractions of the environment, as shown in Figure \ref{fig:worldQuadtree}.
Specifically, each $\mathcal T_q \in \mathcal T^{\mathcal Q}$, $\mathcal T_q \neq \mathcal T_{\mathcal W}$, specifies a precise relation between the leaf nodes of $\mathcal T_{\mathcal W}$ and the leaf nodes of $\mathcal T_q$, an example of which is shown in Figures \ref{fig:worldQuadtree} and \ref{fig:gridAndQtreeAbstraction}.
That is, the tree $\mathcal T_q \in \mathcal T^{\mathcal Q}$ specifies an abstraction for which the leaf nodes of $\mathcal T_{\mathcal W}$ are mapped to leaf nodes of $\mathcal T_q$ in such a way that $\mathcal T_q$ is a pruned quadtree representation of $\mathcal W$.
An alternative way to view this is to consider each $\mathcal T_q \in \mathcal T^{\mathcal Q}$ as a pruned version of $\mathcal T_{\mathcal W}$, where some nodes in the interior of $\mathcal T_{\mathcal W}$ are leaf nodes of $\mathcal T_q$.
In this way, we can consider each $\mathcal T_q \in \mathcal T^{\mathcal Q}$ as encoding an abstraction, or compression, of $\mathcal W$ with a constraint that $\mathcal T_q$ be a valid quadtree depiction of $\mathcal W$.
\begin{figure*}
\begin{minipage}{0.5\textwidth}
	\centering
	\begin{adjustbox}{max size={0.9\textwidth}}
		\begin{tikzpicture}[level distance=1.2cm,
		level 1/.style={sibling distance=3cm},
		level 2/.style={sibling distance=0.7cm}]
		
		\node at (0,0.4) {$\mathcal T_{\mathcal W}$};
		
		\node {$\diamond$}
		child {node {$\diamond$}
			child {node {$x_1$}}
			child {node {$x_2$}}
			child {node {$x_5$}}
			child {node {$x_6$}}
		}
		child {node {$\diamond$}
			child {node {$x_3$}}
			child {node {$x_4$}}
			child {node {$x_7$}}
			child {node {$x_8$}}
		}
		child {node {$\diamond$}
			child {node {$x_9$}}
			child {node {$x_{10}$}}
			child {node {$x_{13}$}}
			child {node {$x_{14}$}}
		}
		child {node {$\diamond$}
			child {node {$x_{11}$}}
			child {node {$x_{12}$}}
			child {node {$x_{15}$}}
			child {node {$x_{16}$}}
		};
		\end{tikzpicture}
		
	\end{adjustbox}
	\\
	\vspace{0.3cm}
		\begin{adjustbox}{max size={0.7\textwidth}}
		\begin{tikzpicture}[scale=.8,every node/.style={minimum size=1cm}]
		
		\begin{scope}[
		yshift=-120,every node/.append style={
			yslant=0,xslant=0},yslant=0,xslant=0
		]
		\fill[white,fill opacity=0.9] (0,0) rectangle (4,4);
		\draw[step=1cm, black] (0,0) grid (4,4); 
		
		\draw[black,very thick] (0,0) rectangle (4,4);
		
		\node at (0.5,0.5) {$x_1$};
		\node at (1.5,0.5) {$x_2$};
		\node at (2.5,0.5) {$x_3$};
		\node at (3.5,0.5) {$x_4$};
		\node at (0.5,1.5) {$x_5$};
		\node at (1.5,1.5) {$x_6$};
		\node at (2.5,1.5) {$x_7$};
		\node at (3.5,1.5) {$x_8$};
		\node at (0.5,2.5) {$x_9$};
		\node at (1.5,2.5) {$x_{10}$};
		\node at (2.5,2.5) {$x_{11}$};
		\node at (3.5,2.5) {$x_{12}$};
		\node at (0.5,3.5) {$x_{13}$};
		\node at (1.5,3.5) {$x_{14}$};
		\node at (2.5,3.5) {$x_{15}$};
		\node at (3.5,3.5) {$x_{16}$};
		
		\end{scope}

		to[out=0,in=120] (-0.2,-3.5);
		\end{tikzpicture}
	\end{adjustbox}
	\caption{Representation of the tree $\mathcal T_{\mathcal W}$ and corresponding grid for a $4 \times 4$ environment.}
	\label{fig:worldQuadtree}
\end{minipage}
~~
\begin{minipage}{0.5\textwidth}
	\centering
	\begin{adjustbox}{max size={.9\textwidth}}
		\begin{tikzpicture}[level distance=1.2cm,
		level 1/.style={sibling distance=3cm},
		level 2/.style={sibling distance=0.7cm}]
		
		\node at (0,0.4) {$\mathcal T_q$};
		
		\node {$\diamond$}
		child {node {$\diamond$}
			child {node {$t_1$}}
			child {node {$t_2$}}
			child {node {$t_3$}}
			child {node {$t_4$}}
		}
		child {node {$t_5$}
		}
		child {node {$t_6$}
		}
		child {node {$\diamond$}
			child {node {{$t_7$}}}
			child {node {$t_8$}}
			child {node {$t_{9}$}}
			child {node {$t_{10}$}}
		};
		\end{tikzpicture}
	\end{adjustbox}
	\\
	\vspace{0.3cm}
	\centering
	\begin{adjustbox}{max size={0.7\textwidth}}
		\begin{tikzpicture}[scale=.8,every node/.style={minimum size=1cm}]
		
		\begin{scope}[
		yshift=-120,every node/.append style={
			yslant=0.0,xslant=0},yslant=0.0,xslant=0
		]
		\fill[white,fill opacity=0.9] (0,0) rectangle (4,4);
		\draw[step=1cm, black] (0,0) grid (4,4); 
		
		\fill[white] (0,2) rectangle (2,4);
		\fill[white] (2,0) rectangle (4,2);
		\draw[black,very thick] (0,0) rectangle (4,4);
		
		\draw[black] (0,2) rectangle (2,4);
		\draw[black] (2,0) rectangle (4,2);

		\draw[black] (0,2) rectangle (2,4);
		\draw[black] (2,0) rectangle (4,2);
		
		\node at (0.5,0.5) {$t_1$};
		\node at (1.5,0.5) {$t_2$};
		
		\node at (0.5,1.5) {$t_3$};
		\node at (1.5,1.5) {$t_4$};
		
		\node at (2.5,2.5) {$t_7$};
		\node at (3.5,2.5) {$t_8$};
		
		\node at (2.5,3.5) {$t_9$};
		\node at (3.5,3.5) {$t_{10}$};
		
		\node at (3,1) {$t_5$};
		\node at (1,3) {$t_6$};
		\end{scope}
		
		to[out=0,in=120] (-0.2,-3.5);
		\end{tikzpicture}
	\end{adjustbox}
	
	\caption{Representation of some $\mathcal T_{q} \in \mathcal T^{\mathcal Q}$ and corresponding grid for a $4 \times 4$ environment.}

	\label{fig:gridAndQtreeAbstraction}
\end{minipage}
\end{figure*}


Per the above discussion, varying the abstraction granularity of $\mathcal W$ can be equivalently viewed as selecting various trees $\mathcal T_q$ in the space  $\mathcal T^{\mathcal Q}$.
Our problem is then one of selecting a tree $\mathcal T_{q} \in \mathcal T^{\mathcal Q}$ as a function of the agent's computational capabilities.
 
The observation that each $\mathcal T_q \in \mathcal T^{\mathcal Q}$ encodes a compression of $\mathcal W$ connects our approach to information-theoretic frameworks that consider optimal encoder design.
The optimization problem to obtain optimal encoders  has been extensively studied by information theorists in the more general setting of signal compression, where no specific structure on the abstraction is enforced (i.e., the resulting encoding need not correspond to any tree representation).
As such, the added constraint that our abstraction be a valid quadtree representation of $\mathcal W$ creates additional challenges, since direct application of information-theoretic methods is not possible.
Thus, to elucidate the technical aspects of our approach, we first present a brief review of the necessary information-theoretical concepts which we will utilize in the formulation of our problem.

\subsection{Information-Theoretical Signal Compression}

The task of obtaining optimal compressed representations of signals is addressed within the realm of information theory \cite{Tishby1999,Slonim2000,Slonim2002,Strouse2017,Cover2006,Hassanpour2017}.
Let $\left ( \Omega,\mathcal F, \mathbb P \right)$ to be a probability space with finite sample space $\Omega$, $\sigma$-algebra $\mathcal F$ and probability measure $\mathbb P: \mathcal F \to \left[0,1\right]$, and denote the set of real and positive real numbers as $\Re$ and $\Re_{++}$, respectively. 
Let $X: \Omega \to \Re$ denote the random variable corresponding to the original, uncompressed, signal, where $X$ takes values in the set $\Omega_X = \left\{x \in \Re : X(\omega) = x, ~\omega \in \Omega \right\}$ and, for any $x \in \Re$, $\p(x) = \mathbb{P}(\left\{\omega \in \Omega : X(\omega) = x \right\})$.
Furthermore, let the random variable $T: \Omega \to \Re$ denote the compressed representation of $X$, where $T$ takes values in the set $\Omega_T = \left\{t \in \Re : T(\omega) = t,~\omega \in \Omega \right\}$.
The level of compression between random variables $X$ and $T$ is measured by the mutual information \cite{Tishby1999,Cover2006}, given by 
\begin{equation} \label{eq:mutual_info_def}
\I(T;X) \triangleq \sum_{t,x} \p(t,x) \log \frac{\p(t,x)}{\p(t) \p(x)}.
\end{equation}
The goal is then to find a stochastic mapping (encoder), denoted $p(t|x)$, which maps outcomes in the uncompressed space $x \in \Omega_X$, to outcomes in the compressed representation $t \in \Omega_T$ so as to minimize $I(T;X)$ (maximize compression) \cite{Tishby1999}.
However, in order to obtain non-trivial solutions, a metric quantifying the quality of the resulting compression must be introduced, since maximal compression $\left( I(T;X) = 0 \right)$ is always achievable. 
The information bottleneck (IB) method~\cite{Tishby1999} defines the quality of the compression utilizing mutual information.

More specifically, the IB method introduces an additional random variable, $Y:\Omega \to \Re$, taking values in the set $\Omega_Y = \left\{y \in \Re: Y(\omega) = y, ~\omega \in \Omega \right\}$. 
The variable $Y$ represents information we are interested in preserving when forming the compressed representation $T$~\cite{Tishby1999,Slonim2000}.
The method imposes the Markov chain condition $T \leftrightarrow  X \leftrightarrow Y$ which arises as a consequence of the problem formulation.
To see this, note that $p(y|t,x) = p(y|x)$ since it is not possible for $t$ to convey any additional information regarding $y$ than what is already in $x$, and thus $T \rightarrow X \rightarrow Y$.
Furthermore, if $p(y|t,x) = p(y|x)$ then $p(t|y,x) = p(t|x)$ which gives $Y \rightarrow X \rightarrow T$.
Therefore, $T \rightarrow X \rightarrow Y$ implies $Y \rightarrow X \rightarrow T$, which is written as $T \leftrightarrow  X \leftrightarrow Y$ \cite{Cover2006,Tishby1999}.

The IB problem is then formulated as
\begin{equation} \label{eq:info_bottleneck_min_prob}
\min_{\p(t|x)}~\I(T;X),
\end{equation}
subject to
\begin{equation} \label{eq:info_bottleneck_min_prob_cons}
\I(T;Y) \geq \hat {D},
\end{equation}
where the minimization is done over all normalized distributions $\p(t|x)$ assuming that the joint distribution $\p(x,y)$ is provided and $\hat D \geq 0$ \cite{Tishby1999}.
Through the introduction of a Lagrange multiplier, $\beta \geq 0$, we have that \eqref{eq:info_bottleneck_min_prob} subject to \eqref{eq:info_bottleneck_min_prob_cons} has Lagrangian
\begin{equation} \label{eq:info_bottleneck_lagrangian}
\mathcal K_Y(\p(t|x);\beta) \triangleq \I(T;X) - \beta \I(T;Y).
\end{equation}
Furthermore, for given $\beta \geq 0$, the optimization problem
\begin{equation} \label{eq:info_bottleneck_largrangian_opt}
\min_{\p(t|x)}~\mathcal K_Y(\p(t|x);\beta),
\end{equation}
can be solved analytically, giving rise to a set of self-consistent equations \cite{Tishby1999}.

The self-consistent equations obtained as a solution to \eqref{eq:info_bottleneck_largrangian_opt} can be solved numerically by an algorithm that likens that of the Blahut-Arimoto algorithm from rate-distortion theory, albeit with no guarantee of convergence to a globally optimal solution \cite{Tishby1999}.
The parameter $\beta$ serves the role of adjusting the amount of relevant information regarding $Y$ that is retained in the abstract representation $T$.
As a result, when $\beta \to \infty$ the optimization process is concerned with the maximal preservation of information, while $\beta \to 0$ promotes maximal compression, with no regard to the information carried regarding $Y$. 
Intermediate values of $\beta$ lead to a spectrum of solutions between these two extremes \cite{Tishby1999}.
The mapping $\p^*(t|x)$ obtained as a solution to the IB problem is generally stochastic, resulting in a deterministic mapping only when $\beta \to \infty$ \cite{Strouse2017,Tishby1999}.

\subsection{Agglomerative Information Bottleneck} 

The agglomerative IB (AIB) method is another framework to form compressed representations of $X$, which is useful when deterministic clusters that retain predictive information regarding the relevant variable $Y$ are desired.
The method uses the IB approach to solve for deterministic, or hard, encoders (i.e., $p(t|x) \in \left\{0,1\right\}$ for all $t$, $x$).
Concepts from AIB will prove useful in our formulation, since each tree $\mathcal T_q \in \mathcal T^{\mathcal Q}$ encodes a hard (deterministic) abstraction of $\mathcal W$, where each leaf node of $\mathcal T_{\mathcal W}$ is aggregated to a specific leaf node of $\mathcal T_{q}$.
That is, by viewing the uncompressed space ($\Omega_X$) as the nodes in $\mathcal N_{\text{leaf}}(\mathcal T_{\mathcal W})$ and the abstracted (compressed) space ($\Omega_T$) as the nodes in $\mathcal N_{\text{leaf}}(\mathcal T_q)$, the abstraction operation can be specified in terms of an encoder $p(t|x)$ where $p(t|x) \in \left\{0,1\right\}$ for all $t$ and $x$, where $p(t|x) = 1$ if $x \in \mathcal N_{\text{leaf}}(\mathcal T_{\mathcal W})$ is aggregated to $t \in \mathcal N_{\text{leaf}}(\mathcal T_q)$, and zero otherwise (see Figures \ref{fig:worldQuadtree} and \ref{fig:gridAndQtreeAbstraction}).
To better understand these connections, we briefly review the AIB before presenting the formulation of our problem.

The solution provided by AIB is an encoder $\p(t|x)$ for which $p(t|x) \in \left\{0,1\right\}$ for all $t$, $x$ and $\beta > 0$.
AIB considers the optimization problem 
\begin{equation} \label{eq:maximization_info_bottleneck_lagrangian}
\max_{\p(t|x)} ~\mathcal L_Y(\p(t|x);\beta),
\end{equation}
where the Lagrangian is defined as
\begin{equation} \label{eq:agglomerative_IB_Max_Lagrangian}
\mathcal L_Y (\p(t|x);\beta) \triangleq \I(T;Y) - \frac{1}{\beta} \I(T;X),
\end{equation}
and the maximization is performed over deterministic distributions $\p(t|x)$ for given $\beta > 0$ and $p(x,y)$ \cite{Slonim2000,Slonim2002}.
 
AIB works from bottom-up, starting with $T = X$ and with each consecutive iteration reduces the cardinality of $T$ until $\left|\Omega_T \right| = 1$\cite{Slonim2000}.
Specifically, let $T_m$ represent the abstracted space with $m$ elements $\left(\left| \Omega_{T_m} \right| = m\right)$ and let $T_i$ represent the compressed space with $\left|\Omega_{T_{i}}\right| = i < m$ elements, where $i = m - 1$ and the number of merged elements is $n = 2$.
We then merge elements $\left\{ t'_{1}, \ldots , t'_{n} \right\} \subseteq \Omega_{T_m}$ to a single element $t \in \Omega_{T_i}$ to obtain $T_i$.
The set $\left\{  t'_{1}, \ldots , t'_{n} \right\} \subseteq \Omega_{T_m}$ selected to merge is determined by considering the difference in the IB Lagrangian induced by the merge operation, as follows.
Let $\p^{-}: \Omega_{T_m} \times \Omega_X \to \left\{0,1\right\}$ be the mapping before the merge and $\p^{+}: \Omega_{T_i} \times \Omega_X \to \left\{0,1\right\}$ be the resulting mapping after elements $\left\{  t'_{1}, \ldots , t'_{n} \right\} \subseteq \Omega_{T_m}$ are grouped to $t \in \Omega_{T_i}$.
Note that, as AIB considers a sequence of merges, the mapping $p^{-}(t|x)$ represents an abstraction of higher cardinality as compared to $p^{+}(t|x)$.
The merger cost is then given by $\Delta \mathcal L_Y: 2^{\Omega_{T_m}}\times \Re_{++} \to \Re$, defined as \cite{Slonim2002}
\begin{equation} \label{eq:agglomerative_IB_Method_mergeCost_1}
	\Delta \mathcal L_Y ( \left\{ t'_{1}, \ldots , t'_{n} \right\} ; \beta) \triangleq \mathcal L_Y(p^{-}(t|x);\beta) - \mathcal L_Y (p^{+}(t|x);\beta).
\end{equation}
The above relation can be decomposed into a change in mutual information by utilizing \eqref{eq:agglomerative_IB_Max_Lagrangian} as
\begin{align} \label{eq:delta_L_agg_IB_1}
	\Delta \mathcal L_Y( \left\{ t'_{1}, \ldots , t'_{n} \right \} ;\beta) = \left[\I(T_m;Y)-\I(T_i;Y)\right] - \frac{1}{\beta}\left[\I(T_m;X) - \I(T_i;X)\right],
\end{align}
which can be further simplified by noting that
\begin{equation}
\I(T;X) = \H(T) - \H(T|X) = \H(T),
\end{equation} 
and where, since $\p(t|x) \in \left\{0,1\right\}$, there is no uncertainty in $T$ once we are provided $x \in \Omega_X$ leading to $\H(T|X) = 0$.
Thus, equation \eqref{eq:delta_L_agg_IB_1} becomes
\begin{align} \label{eq:delta_L_agg_2}
	\Delta \mathcal L_Y( \left\{ t'_{1}, \ldots , t'_{n} \right\} ;\beta) = \left[\I(T_m;Y)-\I(T_i;Y)\right] - \frac{1}{\beta}\left[\H(T_m) - \H(T_i)\right].
\end{align}
It was shown in \cite{Slonim2000,Slonim2002} that \eqref{eq:delta_L_agg_2} can be written as 
	\begin{align}
		\Delta \mathcal L_Y ( \left\{ t'_{1}, \ldots , t'_{n} \right\} ;\beta) = \p(t)\left[\js(\p(y|t'_{1}),\ldots,\p(y|t'_{n})) - \frac{1}{\beta} \H(\Pi)\right],\label{eq:delta_L_agg_3}
	\end{align}
	where $\Pi \in \Re^n$ is given as
	\begin{equation}
	\Pi = \left[ \Pi_1,\ldots,\Pi_n \right]^\mathsf{T} \triangleq \left[ \frac{\p(t'_{1})}{\p(t)},\ldots,\frac{\p(t'_{n})}{\p(t)} \right]^\mathsf{T},
	\end{equation}
	and $\js(\p_1,\ldots,\p_n)$ is the Jensen-Shannon (JS) divergence between the distributions $\p_1,\ldots,\p_n$, with weights $\Pi$ defined as \cite{Lin1991}
	\begin{equation} \label{eq:JSwKLDivergence}
	\js(\p_1,\ldots,\p_n) \triangleq \sum_{s=1}^n \Pi_s \dkl(\p_s,\bar \p),
	\end{equation}
	where, for each outcome $y \in \Omega_Y$,
	\begin{equation}
	\bar \p (y) = \sum_{s=1}^n \Pi_s \p_s(y),
	\end{equation}
	with $\dkl(\mu,\nu)$ denoting the Kullback-Leibler (KL) divergence between probability distributions $\mu$ and $\nu$ given by
	\begin{equation}
	\dkl(\mu,\nu) \triangleq \sum_y \mu(y) \log \frac{\mu(y)}{\nu(y)}.
	\end{equation}
Furthermore, we have that 
\begin{align}
\p(t) &= \sum_{s=1}^{n} \p(t'_{s}) \label{eq:agg_info_bottleneck_p_t}, \\
\p(y| t) &= \sum_{s=1}^{n} \Pi_s\p(y|t'_{s}) \label{eq:agg_info_bottleneck_py_t}, 
\end{align} 
which can be found by realizing that $\p(t|x) \in \left\{0,1\right\}$ for all $x\in\Omega_X$ and $t\in\Omega_T$ and $T \leftrightarrow  X \leftrightarrow Y$\cite{Slonim2000,Slonim2002}.
Note that the merger cost \eqref{eq:agglomerative_IB_Method_mergeCost_1} can be written in terms of the distributions $\p(y|t'_{1}),\ldots,\p(y|t'_{n})$ and the 
weight vector $\Pi$.
This reduces the overall complexity of computing $	\Delta \mathcal L_Y ( \left\{ t'_{1}, \ldots , t'_{n} \right\} ; \beta)$ as opposed to utilizing equation \eqref{eq:delta_L_agg_IB_1}, which contains sums over the sample spaces of $Y$, $T$ and $X$ \cite{Slonim2000, Slonim2002}. 


\section{Problem Formulation} \label{sec:Prob_formulation}

The IB methods presented in the previous section do not impose any constraints on the resulting mapping $\p(t|x)$.
That is, by solving the IB problem, one obtains a mapping $\p^*(t|x)$ that is generally stochastic, and thus it is not guaranteed that it  encodes a (quad)tree representation for any value of $\beta > 0$.
The difficulty lies in the specific structure imposed on the abstraction by the space $\mathcal T^{\mathcal Q}$, as even AIB or deterministic IB cannot guarantee that the resulting $p^*(t|x)$ encode a tree belonging to $\mathcal T^{\mathcal Q}$, although they do provide deterministic encoders \cite{Strouse2017, Slonim2000,Slonim2002}.
Recall that, since each $\mathcal T_{\q} \in \mathcal T^{\mathcal{Q}}$ represents an abstraction of $\mathcal T_{\mathcal W}$, $\mathcal T_{q}$ can be equivalently represented as $\p^{\q}(t|x)$, where $\p^{q}(t|x) = 1$ if $x \in \mathcal N_{\text{leaf}}(\mathcal T_{\mathcal W})$ is abstracted to $t \in \mathcal N_{\text{leaf}}(\mathcal T_q)$ and zero otherwise.
We can then define the IB Lagrangian in the space of quadtrees as the mapping $\L_Y: \mathcal T^{\mathcal Q} \times \Re_{++} \to \Re$ given by
\begin{equation} \label{eq:tree_IB_Lagrangian}
\L_Y(\mathcal T_{\q};\beta) \triangleq \mathcal{L}_Y(\p^{\q}(t|x);\beta),
\end{equation}
where $\mathcal{L}_Y(\p(t|x);\beta)$ is defined in \eqref{eq:agglomerative_IB_Max_Lagrangian}.
Then, for a given $\beta > 0$, we can search the space of trees for the one that maximizes \eqref{eq:tree_IB_Lagrangian}.
This optimization problem is formally given by
\begin{equation} \label{eq:IB_tree_optim_prob_1}
\mathcal T_{\q^*} = \hspace{0.15cm}\stackrel[\mathcal T_{\q} \in \mathcal T^{\mathcal Q}]{}{\argmax}\L_Y(\mathcal T_{\q};\beta).
\end{equation}
The resulting world representation is encoded by the mapping $\p^{\q^*}(t|x)$.
That is, the leafs of $\mathcal{T}_{\q^*}$ determine the optimal multi-resolution representation of $\mathcal{W}$ for the given $\beta$.

By posing the optimization problem as in \eqref{eq:IB_tree_optim_prob_1}, we have implicitly incorporated the constraints on the mapping $\p(t|x)$ in order for the resulting representation to be a quadtree depiction of the world.
While the optimization problem given by \eqref{eq:IB_tree_optim_prob_1} allows one to form an analogous problem to that in \eqref{eq:maximization_info_bottleneck_lagrangian} over the space of trees, the drawback of this method is the need to exhaustively enumerate all feasible quadtrees which can represent the space.
In other words, \eqref{eq:IB_tree_optim_prob_1} requires that $\p^{\q}(t|x)$ be provided for each $\mathcal T_{\q} \in \mathcal T^{\mathcal Q}$.
Because of this, the problem becomes intractable for large grid sizes and thus requires reformulation to handle larger world maps.

\begin{figure}[tbh]
		\centering
		\begin{adjustbox}{max size={0.5\textwidth}}
			\begin{tikzpicture}[level distance=1.2cm,
			level 1/.style={sibling distance=3cm},
			level 2/.style={sibling distance=0.7cm}]
			
			\node at (0,0) {$t_0$};        
			\node at (7,0) {$\mathcal T_{\q^0}$};
			
			\node (A) at (0, -0.5) {};
			\node (B) at (0, -1.5) {};
			
			\node (A2) at (7, -0.5) {};
			\node (B2) at (7, -1.5) {};
			
			\draw [->, line width=0.5mm, black] (A) -- (B);
			\draw [->, line width=0.5mm, black] (A2) -- (B2);
			
			\node at (7,-2) {$\mathcal T_{\q^1}$};
		
			\node at (0,-2) {$\diamond$}
			child {node {$t_{1}$}
			}
			child {node {$t_{2}$}
			}
			child {node {$t_{3}$}
			}
			child {node {$t_{4}$}
			};
			
			\node (C) at (0, -2.5) {};
			\node (D) at (0, -3.5) {};
		
			\node (C2) at (7, -2.5) {};
			\node (D2) at (7, -3.5) {};
		
			\draw [->, line width=0.5mm, black] (C) -- (D);
			\draw [->, line width=0.5mm, black] (C2) -- (D2);
			
			\node at (7,-4) {$\mathcal T_{\q^2}$};
		
			\node at (0,-4) {$\diamond$}
			child {node {$\diamond$}
				child {node {$t_{1_1}$}}
				child {node {$t_{1_2}$}}
				child {node {$t_{1_3}$}}
				child {node {$t_{1_4}$}}
			}
			child {node {$t_2$}
			}
			child {node {$t_3$}
			}
			child {node{$t_4$}
			};
			
			\node (E) at (0, -4.5) {};
			\node (F) at (0, -5.5) {};
			
			\node (E2) at (7, -4.5) {};
			\node (F2) at (7, -5.5) {};
			
			\draw [->, line width=0.5mm, black] (E) -- (F);
			\draw [->, line width=0.5mm, black] (E2) -- (F2);
			
			\node at (7,-6) {$\mathcal T_{\q^3}$};
			
			\node at (0, -6){$\diamond$}
			child {node {$\diamond$}
				child {node {$t_{1_1}$}}
				child {node {$t_{1_2}$}}
				child {node {$t_{1_3}$}}
				child {node {$t_{1_4}$}}
			}
			child {node {$t_2$}
			}
			child {node {$t_3$}
			}
			child {node {$\diamond$}
				child {node {$t_{4_1}$}}
				child {node {$t_{4_2}$}}
				child {node {$t_{4_3}$}}
				child {node {$t_{4_4}$}}
			};
			\end{tikzpicture}
		\end{adjustbox}
	\caption{Sequence of trees from $\mathcal T_{\q^0} = \mathrm{Root}(\mathcal T_{\mathcal W})\in \mathcal T^{\mathcal Q}$ to $\mathcal T_{\q^m} \in \mathcal T^{\mathcal Q}$ ($m=3$) by performing only a sequence of nodal expansions.  Note that $\mathcal T_{\q^0} = \mathrm{Root}\left( \mathcal T_{\mathcal W}\right)$ is the root node of $\mathcal T_{\mathcal W}$.}
	\label{fig:sequnce_of_trees_1}
\end{figure}
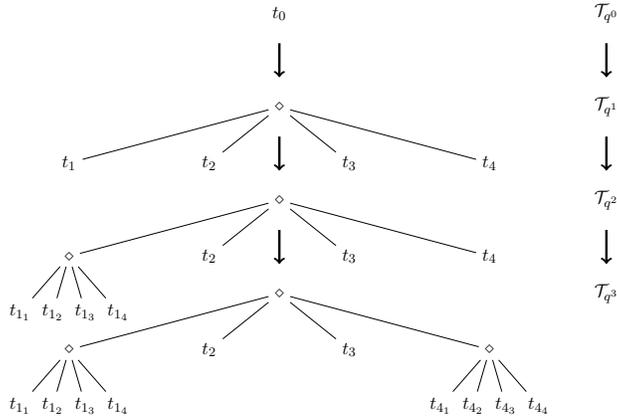

Interestingly,  we note that it is possible to arrive at a quadtree $\mathcal T_{\q^m} \in \mathcal T^{\mathcal{Q}}$ starting from $\mathcal T_{\q^0} \in \mathcal T^{\mathcal Q}$ and performing a sequence of expansions, as illustrated in Figure \ref{fig:sequnce_of_trees_1}.
The resulting sequence of expansions can be viewed as defining a path between $\mathcal T_{\q^0}$ and $\mathcal T_{\q^m}$, in which each vertex of the path corresponds to a distinct intermediate tree in the sequence.
It should be noted that by considering this sequence of expansions it is not always possible to reach any tree $\mathcal T_{\q^m}$ starting from any tree $\mathcal T_{\q^0}$.
In order to address this, we first require the following definitions.
\begin{definition}[\hspace{-0.5pt}\cite{Bundy76}] \label{def:subgraph_subtree}
	\normalfont{
	A tree $\mathcal{G} = (\mathcal{N}(\mathcal G), \mathcal E (\mathcal G))$ is a \emph{subtree} of the 
	tree $\mathcal J = \left(\mathcal N(\mathcal J), \mathcal E(\mathcal J)\right)$, denoted $\mathcal{G} \subseteq \mathcal J$, if $\mathcal{N}(\mathcal G) \subseteq \mathcal N(\mathcal J)$ and $\mathcal E (\mathcal G) \subseteq \mathcal E(\mathcal J)$.
	}
\end{definition}
\begin{definition} \label{def:tree_neighbor_1}
	\normalfont{
	The trees $\mathcal T_{\q'}\in \mathcal T^{\mathcal Q}$ and $\mathcal T_{\q} \in \mathcal T^{\mathcal Q}$ are \emph{neighbors} if $\mathcal{N}(\mathcal T_{\q'}) \setminus \mathcal{N}(\mathcal T_{\q}) = \left\{t'_{1},\ldots,t'_{n} \right\} \subseteq \mathcal N_{\text{leaf}}(\mathcal T_{\q'})$ such that
	$t = \mathcal{P}(t'_{1}) = \cdots = \mathcal{P}(t'_{n}) \in \mathcal N_{\text{leaf}}(\mathcal T_{\q})$ or $\mathcal{N}(\mathcal T_{\q}) =\mathcal{N}(\mathcal T_{\q'}) \setminus \left\{t'_{1},\ldots,t'_{n} \right\}$ where $\left\{t'_{1},\ldots,t'_{n} \right\} \subseteq \mathcal N_{\text{leaf}}(\mathcal T_{\q'})$ have common parent 
	$t = \mathcal{P}(t'_{1}) = \cdots = \mathcal{P}(t'_{n}) \in \mathcal N_{\text{leaf}}(\mathcal T_{\q})$. 
	}
\end{definition}
\vspace{0.1cm}

With these definitions, we see that if $\mathcal T_{\q'} \in \mathcal T^{\mathcal Q}$ is a neighbor of $\mathcal T_{\q} \in \mathcal T^{\mathcal Q}$, then we can obtain $\mathcal T_{\q'}$ by adding the nodes $\left\{t'_{1},\ldots,t'_{n} \right\}$ to $\mathcal T_{\q}$, where the set $\left\{t'_{1},\ldots,t'_{n} \right\}$ consists of the children of a leaf node of $\mathcal T_{\q}$.
We call this process of adding $\left\{t'_{1},\ldots, t'_{n}\right\}$ to $\mathcal N (\mathcal T_{\q})$ a nodal expansion.
We observe that by only performing a sequence of nodal expansions, a path exists between the trees $\mathcal T_{\q^0} \in \mathcal T^{\mathcal Q}$ and $\mathcal T_{\q^m} \in \mathcal T^{\mathcal Q}$ if $\mathcal T_{\q^0}$ is a subtree of $\mathcal T_{\q^m}$ $\left( \mathcal{T}_{\q^0} \subseteq \mathcal{T}_{\q^m} \right)$.
An illustration of nodal expansion is provided in Figure \ref{fig:sequnce_of_trees_1}, where we also note that each tree $\mathcal T_{\q^{i+1}}$ in the sequence is a neighbor to tree $\mathcal T_{\q^{i}}$ with $i \in \left\{0,1,2 \right\}$.

\begin{figure}[t]
	\centering
	\begin{adjustbox}{max size={0.75\textwidth}}
		\begin{tikzpicture}[level distance=1.2cm,
		level 1/.style={sibling distance=3cm},
		level 2/.style={sibling distance=0.7cm}]
		
		\node at (0,0.5) {$\mathcal T_{\q^1}$};
		
		\node at (0,0) {$\diamond$}
		child {node {$t_1$}
		}
		child {node {$t_2$}
		}
		child {node {$t_3$}
		}
		child {node {$t_4$}
		};
		
		\node (C) at (5, 0) {};
		\node (D) at (9, 4) {};
		\draw [->, line width=0.5mm, blue] (C) -- (D);
		
		\node at (15,7) {$\mathcal T_{\q^2}$};
		\node at (15,6.5) {$\diamond$}
		child {node {$\diamond$}
			child {node {$t_{1_1}$}}
			child {node {$t_{1_2}$}}
			child {node {$t_{1_3}$}}
			child {node {$t_{1_4}$}}
		}
		child {node {$t_2$}
		}
		child {node {$t_3$}
		}
		child {node {$t_4$}
		};
		
		\node (E) at (5, -0) {};
		\node (F) at (9, 1) {};
		\draw [->, line width=0.5mm, blue] (E) -- (F);
		
		\node at (15,-3.5) {$\mathcal T_{\q^5}$};
		\node at (15, -4){$\diamond$}
		child {node {$t_1$}
		}
		child {node {$t_2$}
		}
		child {node {$t_3$}
		}
		child {node {$\diamond$}
			child {node {$t_{4_1}$}}
			child {node {$t_{4_2}$}}
			child {node {$t_{4_3}$}}
			child {node {$t_{4_4}$}}
		};
		
		\node (G) at (5, 0) {};
		\node (H) at (9, -1) {};
		\draw [->, line width=0.5mm, blue] (G) -- (H);

		\node at (15,3.5) {$\mathcal T_{\q^3}$};
		\node at (15, 3){$\diamond$}
		child {node {$t_1$}
		}
		child {node {$\diamond$}
			child {node {$t_{2_1}$}}
			child {node {$t_{2_2}$}}
			child {node {$t_{2_3}$}}
			child {node {$t_{2_4}$}}
		}
		child {node {$t_3$}
		}
		child {node {$t_4$}
		};
		
		\node (I) at (5, 0) {};
		\node (J) at (9, -5) {};
		\draw [->, line width=0.5mm, blue] (I) -- (J);

		\node at (15,0) {$\mathcal T_{\q^4}$};
		\node at (15, -0.5){$\diamond$}
		child {node {$t_1$}
		}
		child {node {$t_2$}
		}
		child {node {$\diamond$}
			child {node{$t_{3_1}$}}
			child {node {$t_{3_2}$}}
			child {node {$t_{3_3}$}}
			child {node {$t_{3_4}$}}
		}
		child {node {$t_4$}
		};
		\end{tikzpicture}
	\end{adjustbox}
	\caption{Tree neighbors of $\mathcal T_{\q^1} = \left\{ \mathcal T_{\q^2}, \mathcal T_{\q^3}, \mathcal T_{\q^4}, \mathcal T_{\q^5} \right\}$ that are of higher leaf-node cardinality.}
	\label{fig:qdtree_nbs_1}
\end{figure}
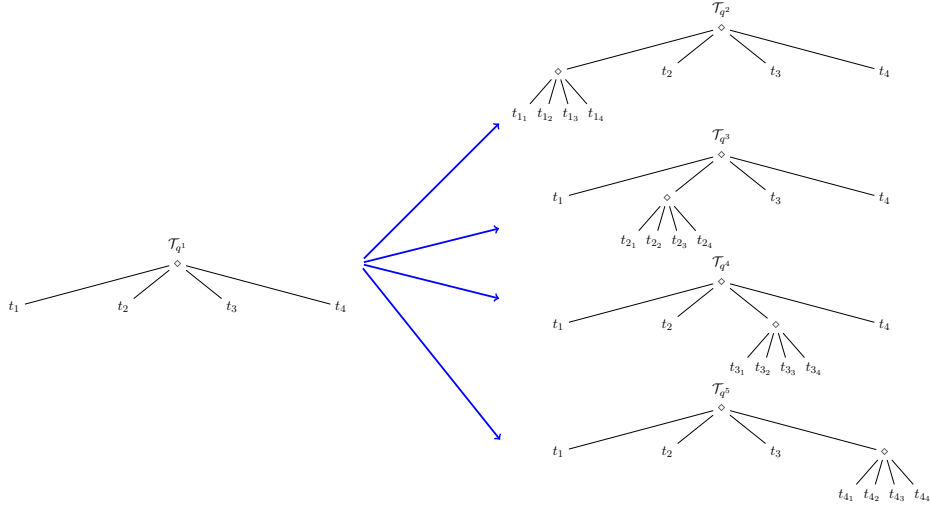

Furthermore, we may view the set of all possible quadtrees as a connected graph, where neighbors are defined according to Definition \ref{def:tree_neighbor_1}.
An illustration of neighboring trees is provided in Figure \ref{fig:qdtree_nbs_1}.
Thus, if it is possible to obtain a sequential characterization of \eqref{eq:tree_IB_Lagrangian}, we can formulate an optimization problem requiring the generation of candidate solutions only along the path leading from $\mathcal T_{\q^0}$ to $\mathcal T_{\q^m}$.
To this end, if we take $\mathcal T_{\q^0} \subseteq \mathcal T_{\q^{m}}$, where $\mathcal T_{\q^0},\mathcal T_{\q^m}\in\mathcal T^{\mathcal{Q}}$, and assume that $\mathcal T_{\q^m}$ is obtained by $m$ expansions of $\mathcal T_{\q^0}$,
then 
\begin{equation} \label{eq:seqCharL}
\L_Y(\mathcal T_{\q^{m}};\beta) = \L_Y(\mathcal T_{\q^0};\beta) + \sum_{i=0}^{m-1} \Delta\L_Y(\mathcal T_{\q^i},\mathcal T_{\q^{i+1}};\beta),
\end{equation}
where $\Delta\L_Y(\cdot,\cdot;\beta)$ is defined as
\begin{equation} \label{eq:tree_delta_L_1}
\Delta\L_Y(\mathcal T_{\q^i},\mathcal T_{\q^{i+1}};\beta) \triangleq \L_Y(\mathcal T_{\q^{i+1}};\beta) - \L_Y(\mathcal T_{\q^i};\beta),
\end{equation}
and $\mathcal T_{\q^{i+1}}\in \mathcal T^{\mathcal Q}$ is a neighbor of $\mathcal T_{\q^{i}} \in \mathcal T^{\mathcal{Q}}$ with higher leaf node cardinality for $i \in \left\{0,\ldots,m-1\right\}$.
Consequently, \eqref{eq:seqCharL} gives a sequential representation of \eqref{eq:tree_IB_Lagrangian}.
Furthermore, the nodal expansion operation to move from tree $\mathcal T_q \in \mathcal T^{\mathcal Q}$ to the neighbor $\mathcal T_{q'} \in \mathcal T^{\mathcal Q}$ has an analogous interpretation to the AIB method discussed in Section \ref{sec:Prelims}.
Consequently,
\begin{equation}
\Delta \L_Y(\mathcal T_{\q^i},\mathcal T_{\q^{i+1}};\beta) = \Delta \mathcal L_Y ( \left\{ t'_{1}, \ldots , t'_{n} \right\} ;\beta),
\end{equation}
and thus
\begin{align}
\Delta \L_Y(\mathcal T_{\q^i},\mathcal T_{\q^{i+1}};\beta) = \p(t)\left[ \js(\p(y|t'_{1}),\ldots,\p(y|t'_{n})) - \frac{1}{\beta}\H(\Pi) \right]. \label{eq:delta_L_agglomerative_IB}
\end{align}
Importantly, note that the structure of $\Delta \L_Y(\mathcal T_{\q^{i}},\mathcal T_{\q^{i+1}};\beta)$ in \eqref{eq:delta_L_agglomerative_IB} only depends on which leafs nodes of $\mathcal T_{\q^i}$ are expanded, as depicted in Figure \ref{fig:invaraiance_delta_L}.
This implies that $\Delta \L_Y(\mathcal T_{\q^{i}},\mathcal T_{\q^{i+1}};\beta)$ is only a function of the nodes that are to be expanded, and not of the overall configuration of the tree, which greatly simplifies the calculation of $\Delta \L_Y(\mathcal T_{\q^i},\mathcal T_{\q^{i+1}};\beta)$.
It follows that the optimization problem can be reformulated as
\begin{equation} \label{eq:IB_tree_optim_prob_2}
\max_m \max_{ \left\{ \mathcal T_{\q^1},\ldots, \mathcal T_{q^{m}} \right\} }~\L_Y(\mathcal T_{\q^0};\beta) + \sum_{i=0}^{m-1} \Delta\L_Y(\mathcal T_{\q^i},\mathcal T_{\q^{i+1}};\beta).
\end{equation}
In this formulation, the constraint encoding that the resulting representation is a quadtree is handled implicitly by $\mathcal T_{\q^i} \in \mathcal T^{\mathcal{Q}}$.
The additional maximization over $m$ in \eqref{eq:IB_tree_optim_prob_2} appears since the horizon of the problem is not known a-priori and is, instead, a free parameter in the optimization problem.

Next, we propose two algorithms that can be used to solve the problem in \eqref{eq:IB_tree_optim_prob_2}.
Note that, by taking $\mathcal T_{\q^0} =\mathrm{Root}(\mathcal T_{\mathcal W} )\in \mathcal T^{\mathcal Q}$, we can guarantee that a path exists between $\mathcal T_{\q^0} $ and any other $\mathcal T_{\q} \in \mathcal T^{\mathcal Q}$, since, in this case, $\mathcal T_{q^0} \subseteq \mathcal T_{q}$ for all $\mathcal T_q \in \mathcal T^{\mathcal Q}$, as shown in Figure \ref{fig:sequnce_of_trees_1}.

\section{Algorithmic Solutions} \label{sec:algorithms}

In this section, we discuss two novel algorithmic approaches to solve the optimization problem~\eqref{eq:IB_tree_optim_prob_2}.
Specifically, we present two approaches: a Greedy search method, and an algorithm we call Q-tree search.
Proofs of all lemmas, propositions and theorems in this section are provided in the appendix.

\vspace{-0.2cm}
\subsection{A Greedy Approach} \label{sunsec:greedy_alg}
\begin{algorithm}[tb]\label{alg:greedy_tree_search}
	\KwData{$\p(x,y)$, $\beta > 0$}
	
	\KwResult{$\mathcal T_{\q^*}$}
	
	\textbf{Initialize:} $\mathcal T_{\q^0}$, $i \leftarrow 0$, $\textit{Stop\_Flag} \leftarrow \texttt{False}$. 
	
	\While{\textrm{\textbf{not}} Stop\_Flag}{
			
		\For{$\mathcal T_{\q}$ neighbor of $\mathcal T_{\q^i}$}{
		
			\textit{neighbor\_vector} $\leftarrow \Delta \L_Y(\mathcal T_{\q^i},\mathcal T_{\q};\beta)$
		
			}
			
		\If{$\max$ neighbor\_vector $> 0$}{
			
			$b \leftarrow \argmax$ \textit{neighbor\_vector} \\
			$\mathcal T_{\q^{i+1}} \leftarrow$ neighbor $\mathcal T_{q^b}$ of $\mathcal T_{\q^i}$ \\
			$i \leftarrow i+1$
			
			}
		\Else{
			\textit{Stop\_Flag} $\leftarrow \texttt{True}$ \\
			$\mathcal T_{\q^*} \leftarrow \mathcal T_{\q^i}$
		}
	}
	\caption{The Greedy Algorithm.}
\end{algorithm}
A Greedy approach to solve \eqref{eq:IB_tree_optim_prob_2} involves maximizing $\Delta\L_Y(\mathcal T_{\q^i},\mathcal T_{\q^{i+1}};\beta)$ myopically at each step.
That is, provided that $\mathcal T_{q^{i+1}} \in \mathcal T^{\mathcal Q}$ is a neighbor of $\mathcal T_{q^i} \in \mathcal T^{\mathcal Q}$, we consider the next tree $\mathcal T_{\q^{i+1}}$ that maximizes the value of $\Delta \L_Y(\mathcal T_{\q^i},\mathcal T_{\q^{i+1}};\beta)$, and we sequentially keep selecting trees ($\mathcal T_{\q^{i+1}}\to \mathcal T_{\q^{i+2}}\to \ldots$) until no further improvement is possible.
In other words, the Greedy algorithm continues along the current path in the space of trees until it finds a tree $\mathcal T_{\q^i} \in \mathcal T^{\mathcal Q}$ that has no neighbor $\mathcal T_{\q^{i+1}} \in \mathcal T^{\mathcal Q}$ of $\mathcal T_{\q^i}$ such that $\Delta \L_Y(\mathcal T_{\q^i},\mathcal T_{\q^{i+1}};\beta) > 0$.
The process is detailed in Algorithm~\ref{alg:greedy_tree_search}.

The Greedy algorithm is simple to implement and requires little pre-processing.
However, one can construct examples for a given $\beta > 0$ and $\mathcal T_{\q^{i}} \in \mathcal T^{\mathcal Q}$ for which $\Delta \L_Y(\mathcal T_{\q^{i}},\mathcal T_{\q^{i+1}};\beta) < 0$ for all $\mathcal T_{\q^{i+1}} \in \mathcal T^{\mathcal Q}$  that are neighbors of $\mathcal T_{\q^{i}}$, and where there exists at least one neighbor $\mathcal T_{\q^{i+2}} \in \mathcal T^{\mathcal Q}$ of $\mathcal T_{\q^{i+1}}$ such that $\Delta \L_Y(\mathcal T_{\q^{i+1}},\mathcal T_{\q^{i+2}};\beta) > 0$ and  $\Delta \L_Y(\mathcal T_{\q^{i}},\mathcal T_{\q^{i+1}};\beta) + \Delta \L_Y(\mathcal T_{\q^{i+1}},\mathcal T_{\q^{i+2}};\beta) > 0$.
This implies that the Greedy algorithm is not able to further improve the value of \eqref{eq:IB_tree_optim_prob_2} at the current tree $\mathcal T_{\q^{i}}$.
In such a scenario, the algorithm will terminate at the condition $\Delta \L_Y(\mathcal T_{\q^{i}},\mathcal T_{\q^{i+1}};\beta) < 0$, without gaining access to $\Delta \L_Y(\mathcal T_{\q^{i+1}},\mathcal T_{\q^{i+2}};\beta) > 0$.
Since in this scenario $\Delta \L_Y(\mathcal T_{\q^{i}},\mathcal T_{\q^{i+1}};\beta) + \Delta \L_Y(\mathcal T_{\q^{i+1}},\mathcal T_{\q^{i+2}};\beta) > 0$, further improvement of \eqref{eq:IB_tree_optim_prob_2} is possible, but not achievable by the Greedy approach.  
Therefore, while the Greedy algorithm is simple to implement, it does not, in general, find globally optimal solutions.
However, as $\beta \to \infty$, the Greedy algorithm does find a global solution as $\lim_{\beta \to \infty}\Delta \L_Y(\mathcal T_{\q^{i}},\mathcal T_{\q^{i+1}};\beta) \geq 0$ for all $\mathcal T_{\q} \in \mathcal T^{\mathcal Q}$, as seen by the limit of \eqref{eq:delta_L_agglomerative_IB} and non-negativity of the JS-divergence.
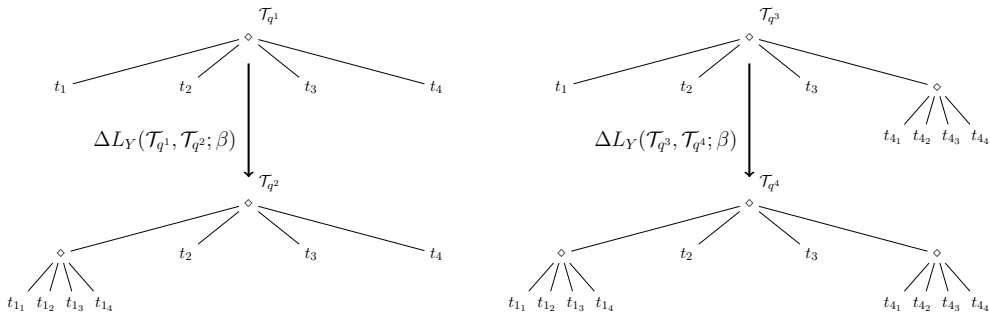
\begin{figure}[b]
	\centering
	\begin{adjustbox}{max size={0.8\textwidth}}
		\begin{tikzpicture}[level distance=1.2cm,
		level 1/.style={sibling distance=3cm},
		level 2/.style={sibling distance=0.7cm}]

	    \node at (-5.5,0.5) {$\mathcal T_{q^1}$};
		\node at (-6,0) {$\diamond$}
		child {node {$t_1$}
		}
		child {node {$t_2$}
		}
		child {node {$t_3$}
		}
		child {node {$t_4$}
		};
		
		\node (C) at (-6, -0.5) {};
		\node (D) at (-6, -3.5) {};
			
		\draw [->, line width=0.5mm, black] (C) -- (D);

		\node at (-5.5,-3.5) {$\mathcal T_{q^2}$};
		\node at (-6,-4) {$\diamond$}
		child {node {$\diamond$}
			child {node {$t_{1_1}$}}
			child {node {$t_{1_2}$}}
			child {node {$t_{1_3}$}}
			child {node {$t_{1_4}$}}
		}
		child {node {$t_2$}
		}
		child {node {$t_3$}
		}
		child {node{$t_4$}
		};
		
		\node (E) at (6, -0.5) {};
		\node (F) at (6, -3.5) {};
		
		\node (L) at (4,-2.5) {\Large{$\Delta \L_Y (\mathcal T_{\q^3},\mathcal T_{\q^4}; \beta)$}};
		\node (L) at (-8,-2.5) {\Large{$\Delta \L_Y (\mathcal T_{\q^1},\mathcal T_{\q^2}; \beta)$}};
			
		\draw [->, line width=0.5mm, black] (E) -- (F);

		\node at (6.5,0.5) {$\mathcal T_{q^3}$};
		\node at (6, 0){$\diamond$}
		child {node {$t_1$}
		}
		child {node {$t_2$}
		}
		child {node {$t_3$}
		}
		child {node {$\diamond$}
			child {node {$t_{4_1}$}}
			child {node {$t_{4_2}$}}
			child {node {$t_{4_3}$}}
			child {node {$t_{4_4}$}}
		};
		
		\node at (6.5,-3.5) {$\mathcal T_{q^4}$};
		\node at (6, -4){$\diamond$}
		child {node {$\diamond$}
			child {node {$t_{1_1}$}}
			child {node {$t_{1_2}$}}
			child {node {$t_{1_3}$}}
			child {node {$t_{1_4}$}}
		}
		child {node {$t_2$}
		}
		child {node {$t_3$}
		}
		child {node {$\diamond$}
			child {node {$t_{4_1}$}}
			child {node {$t_{4_2}$}}
			child {node {$t_{4_3}$}}
			child {node {$t_{4_4}$}}
		};
		\end{tikzpicture}
	\end{adjustbox}
	\caption{Representation of the invariance of $\Delta \L_Y(\mathcal T_{\q^{i}}, \mathcal T_{\q^{i+1}};\beta)$.  In this case, $\Delta \L_Y(\mathcal T_{\q^{1}}, \mathcal T_{\q^2};\beta) = \Delta \L_Y(\mathcal T_{\q^{3}}, \mathcal T_{\q^4};\beta)$.}
	\label{fig:invaraiance_delta_L}
\end{figure}
\vspace{-0.2cm}

\subsection{The Q-tree Search Algorithm}

We now present another approach, detailed in Algorithm~\ref{alg:Q_search}, designed to overcome some of the shortfalls encountered with the Greedy algorithm.
The main drawback by utilizing the Greedy approach in solving the optimization problem~\eqref{eq:IB_tree_optim_prob_2} is the short-sightedness of the algorithm and its inability to realize that poor expansions at the current step may lead to much higher-valued options in the future.
This is analogous to problems in reinforcement learning and dynamic programming, where an action-value function $\left ( Q\text{-function} \right )$ is introduced to incorporate the notion of cost-to-go for selecting among feasible actions in a given state \cite{Bertsekas2012,Sutton1998}.
The idea behind introducing such a function is to incorporate future costs, thus allowing agents to take actions that are not the most optimal with respect to the current one-step cost, but have lower total cost due to events that are possible in the future.

To this end, we define the function
\begin{equation} \label{eq:Q_Y_future}
\Q_{Y}(\mathcal T_{\q^i},\mathcal T_{\q^{i+1}};\beta) \triangleq 
\max \left\{\Delta \L_Y(\mathcal T_{\q^i},\mathcal T_{\q^{i+1}};\beta) + \sum_{\tau=1}^{n} \Q_{Y}(\mathcal T_{\q^{i+1}},\mathcal T_{\q_\tau^{i+2}};\beta)  , ~0 \right\},
\end{equation}
where $\mathcal T_{\q^{i+1}}$ is a neighbor of $\mathcal T_{\q^{i}}$ with higher leaf node cardinality and
\begin{equation}
    Q_Y(\mathcal T_{q'},\mathcal T_{\mathcal W};\beta) \triangleq \max \left\{\Delta L_Y(\mathcal T_{q'},\mathcal T_{\mathcal W};\beta),~0 \right\},
\end{equation}
for all $\mathcal T_{q'}\in \mathcal T^{\mathcal Q}$ for which $\mathcal T_{\mathcal W}\in\mathcal T^{\mathcal Q}$ is a neighbor.
Hence, there exists a $t \in \mathcal{N}_{\text{leaf}}(\mathcal T_{\q^{i}})$ for which $\mathcal{C}(t)=\left\{ t'_{1},\ldots,t'_{n} \right\} = \mathcal{N}(\mathcal T_{\q^{i+1}})\setminus \mathcal{N}(\mathcal T_{\q^{i}})$. 
The quadtrees  $\mathcal T_{\q_\tau^{i+2}},~\tau \in \left\{1,\ldots,n\right\}$ are neighbors of $\mathcal T_{\q^{i+1}}$ which are obtained by expanding the 
leaf nodes $ t'_{\tau} \in \mathcal{C}(t)$ for $\tau = 1,\ldots,n$, as shown in Figure~\ref{fig:delta_L_and_Q}.

\begin{figure}[t]
	\centering
	\begin{adjustbox}{max size={0.8\textwidth}}
		\begin{tikzpicture}[level distance=1.2cm,
		level 1/.style={sibling distance=3cm},
		level 2/.style={sibling distance=0.7cm}]
		
		\node at (0,4) {$t_0$};

		
		\node at (2.2, 1.75){\Large $\Delta \L_Y(\mathcal T_{\q^i},\mathcal T_{\q^{i+1}};\beta)$ \normalsize};
		
		\node (A) at (0, 3) {};
		\node (B) at (0, 0.5) {};
		\draw [->,line width=0.5mm, black] (A) -- (B);
		
		\node at (0,0) {$\diamond$}
		child {node {$t_1$}
		}
		child {node {$t_2$}
		}
		child {node {$t_3$}
		}
		child {node {$t_4$}
		};
		
		\node (C) at (0, -2) {};
		\node (D) at (-7.5, -3.5) {};
		\draw [->, line width=0.5mm, black] (C) -- (D);
		
		\node at (-10, -3.6){\Large $\Q_Y(\mathcal T_{\q^{i+1}},\mathcal T_{\q_{1}^{i+2}};\beta)$ \normalsize};
		\node at (10, -3.6){\Large $\Q_Y(\mathcal T_{\q^{i+1}},\mathcal T_{\q_{4}^{i+2}};\beta)$ \normalsize};
		
		\node at (-8,-4) {$\diamond$}
		child {node {$\diamond$}
			child {node {$t_{1_1}$}}
			child {node {$t_{1_2}$}}
			child {node {$t_{1_3}$}}
			child {node {$t_{1_4}$}}
		}
		child {node {$t_2$}
		}
		child {node {$t_3$}
		}
		child {node {$t_4$}
		};
		
		\node (E) at (0, -2) {};
		\node (F) at (7.5, -3.5) {};
		\draw [->, line width=0.5mm, black] (E) -- (F);
		
		\node at (8, -4){$\diamond$}
		child {node {$t_1$}
		}
		child {node {$t_2$}
		}
		child {node {$t_3$}
		}
		child {node {$\diamond$}
			child {node {$t_{4_1}$}}
			child {node {$t_{4_2}$}}
			child {node {$t_{4_3}$}}
			child {node {$t_{4_4}$}}
		};
		
		\node (G) at (0, -2) {};
		\node (H) at (4, -7) {};
		\draw [->, line width=0.5mm, black] (G) -- (H);
		
		\node at (-8, -10){\Large $\Q_Y(\mathcal T_{\q^{i+1}},\mathcal T_{\q_{2}^{i+2}};\beta)$ \normalsize};
		\node at (8, -10){\Large $\Q_Y(\mathcal T_{\q^{i+1}},\mathcal T_{\q_{3}^{i+2}};\beta)$ \normalsize};
		
		\node at (-8, -7){$\diamond$}
		child {node {$t_1$}
		}
		child {node {$\diamond$}
			child {node {$t_{2_1}$}}
			child {node {$t_{2_2}$}}
			child {node {$t_{2_3}$}}
			child {node {$t_{2_4}$}}
		}
		child {node {$t_3$}
		}
		child {node {$t_4$}
		};
		
		\node (I) at (0, -2) {};
		\node (J) at (-4, -7) {};
		\draw [->, line width=0.5mm, black] (I) -- (J);

		\node at (8, -7){$\diamond$}
		child {node {$t_1$}
		}
		child {node {$t_2$}
		}
		child {node {$\diamond$}
			child {node{$t_{3_1}$}}
			child {node {$t_{3_2}$}}
			child {node {$t_{3_3}$}}
			child {node {$t_{3_4}$}}
		}
		child {node {$t_4$}
		};
		\end{tikzpicture}
	\end{adjustbox}
	\caption{Illustration of $\Q_Y(\mathcal T_{\q^i},\mathcal T_{\q^{i+1}};\beta)$ and its dependency on $\Q_Y(\mathcal T_{\q^{i+1}},\mathcal T_{\q_\tau^{i+2}};\beta)$. 
	Consider that the algorithm is at tree $\mathcal T_{\q^0}$, represented by the single node $t_0$.  
	Each of the nodes $\left\{t'_{1},t'_{2},t'_{3},t'_{4} \right\} = \left\{t_1,t_2,t_3,t_4 \right\}$, which are children of $t_0$, are  expanded one by one to form the trees $\mathcal T_{\q_\tau^{i+2}}$ for $\tau \in \left\{1,2,3,4 \right\}$.  Note that $n = 4$ in \eqref{eq:Q_Y_future} for the special case of quadtrees.}
	\label{fig:delta_L_and_Q}
\end{figure}
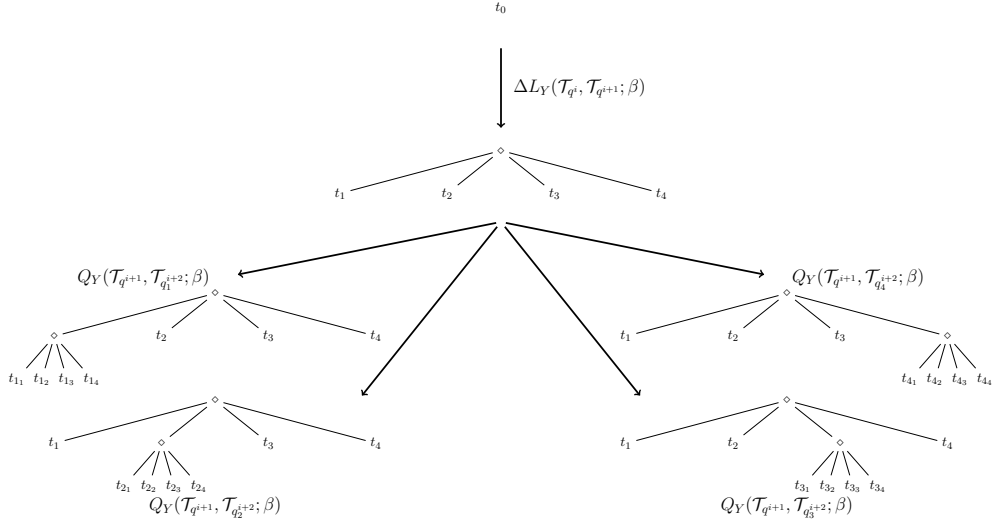
\begin{algorithm}[tb] \label{alg:Q_search}
	\KwData{$\p(x,y)$, $\beta > 0$}
	
	\KwResult{$\mathcal T_{\q^*}$}
	
	\textbf{Initialize:} $\mathcal T_{\q^0}$, $i \leftarrow 0$, $\textit{Stop\_Flag} \leftarrow \texttt{False}$, Populate $\Q_Y(\mathcal T_{\q^i},\mathcal T_{\q^{i+1}};\beta)$. 
	
	\While{\textrm{\textbf{not}} Stop\_Flag}{
		
		\For{$\mathcal T_{\q}$ neighbor of $\mathcal T_{\q^i}$}{
			
			\textit{neighbor\_vector} $\leftarrow \Q_Y(\mathcal T_{\q^i},\mathcal T_{\q};\beta)$
			
		}
		
		\If{$\max$ neighbor\_vector $> 0$}{
			
			$b \leftarrow \argmax$ \textit{neighbor\_vector} \\
			$\mathcal T_{\q^{i+1}} \leftarrow$ neighbor $\mathcal T_{q^b}$ of $\mathcal T_{\q^i}$ \\
			$i \leftarrow i+1$
			
		}
		\Else{
			\textit{Stop\_Flag} $\leftarrow \texttt{True}$ \\
			$\mathcal T_{\q^*} \leftarrow \mathcal T_{\q^i}$
		}
	}
	\caption{The Q-tree search Algorithm.}
\end{algorithm}
Note that $\Q_Y(\mathcal T_{\q^i},\mathcal T_{\q^{i+1}};\beta)$ conveys whether or not a current poor expansion (that is, one where $\Delta \L_Y(\mathcal T_{\q^i},\mathcal T_{\q^{i+1}};\beta) < 0$) can be overcome by future rewards by continuing expansions that are available through $\left\{ t'_{1},\ldots,t'_{n} \right\}$.
Observe that this is possible due to the dependence of $\Delta \L_Y(\mathcal T_{\q^i}, \mathcal T_{\q^{i+1}};\beta)$ on only the nodes added by moving from $\mathcal T_{\q^i}$ to $\mathcal T_{\q^{i+1}}$ and not the overall configuration of the tree, as seen in \eqref{eq:delta_L_agglomerative_IB} and subsequent discussion.
Furthermore, the sum over $\tau$ in \eqref{eq:Q_Y_future} encodes the fact that it is possible for all children of $\left\{ t'_{1},\ldots,t'_{n} \right\}$ to be expanded in ensuing steps if they improve the quality of the solution.
Furthermore, from the definition of $\Q_Y(\mathcal T_{\q^i},\mathcal T_{\q^{i+1}};\beta)$, we see that even if $\sum_{\tau=1}^{n} \Q_{Y}(\mathcal T_{\q^{i+1}},\mathcal T_{\q_\tau^{i+2}};\beta) = 0$ then the algorithm will not ignore a one-step improvement if $\Delta \L_Y(\mathcal T_{\q^i},\mathcal T_{\q^{i+1}};\beta) > 0$. 
In general, the solution obtained by the Greedy algorithm will not necessarily be the same as the one obtained by the Q-tree search algorithm. 
Contrasting the Q-tree search algorithm to the Greedy approach, we obtain the following theorem that relates the solutions obtained by these two methods.

\begin{theorem} \label{thm:Q_search_and_greedy_soln}
	\normalfont{
    Let $\mathcal T_{q^0} \in \mathcal T^{\mathcal Q}$ be a tree at which both Greedy and Q-tree search algorithms are initialized. 
	Then the solution $\mathcal T _{\q^*_\text{G}}$ obtained by the Greedy algorithm is a subtree of the solution $\mathcal T _{\q^*_\text{Q}}$ obtained by the Q-search method.
	}
\end{theorem}
\noindent As a direct consequence of Theorem \ref{thm:Q_search_and_greedy_soln}, solutions obtained by the Q-tree search algorithm 
will contain at least as many leaf-nodes as the solution of the Greedy approach, and, at the same time, produce a better solution (if one exists) with respect to \eqref{eq:IB_tree_optim_prob_2} for a given $\beta > 0$.

Before we discuss the properties of the solution obtained by the Q-tree search algorithm, we provide the following definition of a minimal tree.
\begin{definition} \label{def:minimal}
	\normalfont{
	A tree $\mathcal T_q \in \mathcal T^{\mathcal Q}$ is \emph{minimal} with respect to the cost $L_Y(\cdot;\beta)$ if, for all $\mathcal T_{q'}\in \mathcal T^{\mathcal Q}$ such that $\mathcal T_{q'} \subset \mathcal T_{q}$,  $L_Y(\mathcal T_{q'}; \beta) < L_Y(\mathcal T_{q};\beta)$.
	}
\end{definition} 

From Definition \ref{def:minimal} we see that, if a tree is minimal, then it is not possible to reduce the number of leaf nodes of the tree without reducing the value of the objective function $L_Y(\cdot;\beta)$. 
In what follows, we will show that the tree obtained by the Q-tree search algorithm is minimal and optimal with respect to \eqref{eq:IB_tree_optim_prob_2}.
In order to present these theoretical results, some additional definitions are required, which are provided next.

\begin{figure}[htb]
	\centering
	\begin{adjustbox}{max size={0.7\textwidth}}
		\begin{tikzpicture}[level distance=1.2cm,
		level 1/.style={sibling distance=3.5cm},
		level 2/.style={sibling distance=0.7cm},
		level 3/.style={sibling distance=0.4cm},
		level 4/.style={sibling distance=0.2cm}]
		\node[black!30!] at (0,0.3){$\mathcal T_{q}$};
		\node[black!30!] at (0, 0){$\diamond$} 
		child[black!30!] {node[black!30!] {$\diamond$} 
			child {node[black!30!]{$\diamond$}
				child {node[black!30!]{$\diamond$}}
				child {node[black!30!] {$\diamond$}}
				child {node[black!30!] {$\diamond$}}
				child {node[black!30!] {$\diamond$}}
			}
			child {node[black!30!] {$\diamond$}}
			child {node[black!30!] {$\diamond$}}
			child {node[black!30!] {$\diamond$}
				child {node[black!30!] {$\diamond$}}
				child {node[black!30!] {$\diamond$}}
				child {node[black!30!] {$\diamond$}}
				child {node[black!30!] {$\diamond$}}
			}
		}
		child[black!30!] {node[black!30!] {$\diamond$}}  
		child[black!30!] {node[black!30!] {$\diamond$} 	
			child {node[black!30!]{$\diamond$}}
			child {node[black!100!] {$t$}
				child[black!100!] {node[black!100!] {$t'_1$}
					child[black!100!] {node[black!100!] {$\diamond$}}
					child[black!100!] {node[black!100!] {$\diamond$}}
					child[black!100!] {node[black!100!] {$\diamond$}}
					child[black!100!] {node[black!100!] {$\diamond$}}
				}
				child[black!100!] {node[black!100!] {$t'_2$}}
				child[black!100!] {node[black!100!] {$t'_3$}}
				child[black!100!] {node[black!100!] {$t'_4$}
					child[black!100!] {node[black!100!] {$\diamond$}}
					child[black!100!] {node[black!100!] {$\diamond$}}
					child[black!100!] {node[black!100!] {$\diamond$}}
					child[black!100!] {node[black!100!] {$\diamond$}}
				}
			}
			child {node[black!30!] {$\diamond$}}
			child {node[black!30!] {$\diamond$}}
		}
		child[black!30!] {node[black!30!] {$\diamond$}  
			child {node[black!30!] {$\diamond$}}
			child {node[black!30!] {$\diamond$}
				child {node[black!30!] {$\diamond$}}
				child {node[black!30!] {$\diamond$}}
				child {node[black!30!] {$\diamond$}}
				child {node[black!30!] {$\diamond$}}
			}
			child {node[black!30!] {$\diamond$}}
			child {node[black!30!] {$\diamond$}}
		};
		\end{tikzpicture}
	\end{adjustbox}
	\caption{Visual representation of $\mathcal T_{q(t)}$ (black), where  $\mathcal T_{q(t)} \subseteq \mathcal T_{q}$ for some $\mathcal T_q \in \mathcal T^{\mathcal Q}$ and node $t \in \mathcal N(\mathcal T_q)$.  The children of node $t\in \mathcal N\left(\mathcal T_q\right)$, given by $\mathcal C(t) = \left\{t'_1,t'_2,t'_3,t'_4\right\}$, are also shown.}
	\label{fig:treeRootedAtNodet}
\end{figure}
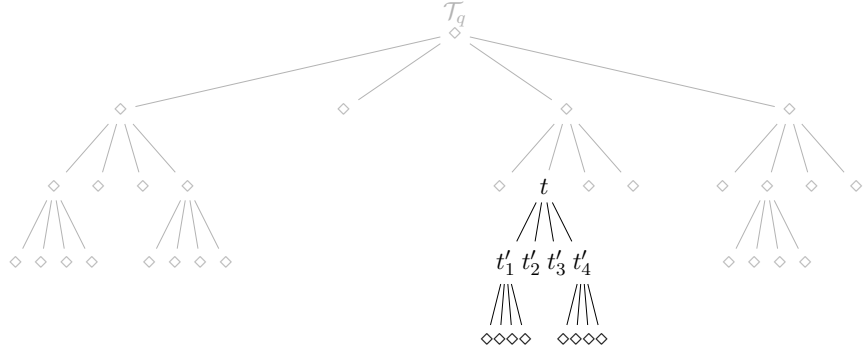

\begin{definition} \label{def:subtree_rooted}
	\normalfont{
	Given any node $t \in \mathcal N(\mathcal T_{q})$, the \emph{subtree} of $\mathcal T_{q} \in \mathcal T^{\mathcal Q}$ \emph{rooted at node} $t$ is denoted by $\mathcal T_{q(t)}$ and has node set
	\begin{equation*}
	\mathcal N\left(\mathcal T_{q(t)} \right) = \Big\{ t' \in \mathcal N(\mathcal T_{q}): t' \in  \bigcup_{i} \mathcal D_i \Big\},
	\end{equation*}
	where $\mathcal D_1 = \left\{ t \right\}$, $\mathcal D_{i+1} = \mathcal A \left( \mathcal D_i \right)$ and where 
	\begin{equation*}
	\mathcal A\left( \mathcal D_i \right) = \Big\{ t' \in \mathcal N(\mathcal T_{\mathcal W}): t' \in \bigcup_{m\in \mathcal{D}_i}\mathcal C\left( m \right) \Big\}.
	\end{equation*}
	}
\end{definition}
\noindent A visualization of $\mathcal T_{q(t)}$ for some $\mathcal T_q \in \mathcal T^{\mathcal Q}$ is provided in Figure \ref{fig:treeRootedAtNodet}.
Furthermore, recall that $\Delta \L_Y(\mathcal T_{\q},\mathcal T_{\q'};\beta)$ is only a function of the nodes that are added to tree $\mathcal T_{q} \in \mathcal T^{\mathcal Q}$ to obtain $\mathcal T_{q'} \in \mathcal T^{\mathcal Q}$, as shown by \eqref{eq:delta_L_agglomerative_IB} and depicted in Figure \ref{fig:invaraiance_delta_L}.
Thus, it is convenient to describe $\Delta \L_Y(\mathcal T_{\q},\mathcal T_{\q'};\beta)$ explicitly as a function of the nodes of the trees $\mathcal T_{q}$ and $\mathcal T_{q'}$ as given in the following definition.
\begin{definition} \label{def:deltaLhat}
	\normalfont{
	 The \emph{node-wise} $\Delta \hat L_Y$-\emph{function} for any node $t\in \mathcal{N}_{\text{int}}(\mathcal{T}_\mathcal{W})$ is given by
	\begin{equation*}
	\Delta \hat L_Y(t;\beta) = \Delta \mathcal L_Y(\left\{ t'_{1} ,\ldots, t'_{n} \right\};\beta ),
	\end{equation*}
	where  $\left\{t'_{1},\ldots,t'_{n} \right\}  = \mathcal C(t) \subset \mathcal N(\mathcal T_{\mathcal{W}})$. 
	Furthermore, $\Delta \hat L(t;\beta) = 0$ for all $t \in \mathcal N_{\text{leaf}}\left(\mathcal T_{\mathcal W}\right)$.
	}
\end{definition}
\noindent As a consequence of Definition \ref{def:deltaLhat}, note that if we let $\mathcal T_{q'} $ be a neighbor of $\mathcal T_{q} $ such that $\left\{t'_{1},\ldots,t'_{n} \right\} = \mathcal C(t) =  \mathcal N(\mathcal T_{q'})\setminus  \mathcal N(\mathcal T_{q})$ where $t \in \mathcal{N}_{\text{leaf}}(\mathcal T_{q})$ then,
\begin{equation}
\Delta L_Y(\mathcal T_{q}, \mathcal T_{q'};\beta) = \Delta \hat L_Y(t;\beta).
\end{equation}
Moreover, since $Q_Y(\cdot,\cdot;\beta)$ in \eqref{eq:Q_Y_future} is recursively defined in terms of $\Delta L_Y(\cdot,\cdot;\beta)$, we have the following definition.
\begin{definition} \label{def:node_wise_Q}
	\normalfont{
	The \emph{node-wise} $\hat Q_Y$-\emph{function} for  any node $t\in \mathcal{N}_{\text{int}}(\mathcal{T}_\mathcal{W})$ is given by  
	\begin{align*}
	\hat Q_Y(t;\beta) = \max \Big\{ \Delta \hat L_Y(t;\beta ) + \sum_{t' \in \mathcal C (t)} \hat Q_Y(t';\beta), ~0\Big\}, 
	\end{align*}
	and where $\hat Q_Y(t;\beta)=0$ for all $t\in \mathcal{N}_{\text{leaf}}(\mathcal{T}_\mathcal{W})$.\\
	}
\end{definition}
\noindent From Definition~\ref{def:node_wise_Q}, if $\mathcal T_{q'}\in \mathcal T^{\mathcal Q}$ is a neighbor of $\mathcal T_{q}\in \mathcal T^{\mathcal Q} $
where nodes $\left\{t'_{1},\ldots,t'_{n} \right\} = \mathcal C(t) \subseteq \mathcal N_{\text{leaf}}(\mathcal T_{q'})$ are merged to a node $t \in \mathcal N_{\text{leaf}}(\mathcal T_{q})$ to obtain tree $\mathcal T_{q}$, then we have
\begin{equation}
Q_Y(\mathcal T_{q},\mathcal T_{q'};\beta) = \hat Q_Y(t;\beta).
\end{equation}
As a result of Definitions~\ref{def:deltaLhat} and \ref{def:node_wise_Q}, if $\left\{ \mathcal T_{q^i},\mathcal T_{q^{i+1}},\ldots,\mathcal T_{q^{i+j}} \right\} \subset \mathcal T^{\mathcal Q}$ is a sequence of trees such that $\mathcal T_{q^{i+k+1}}$ is a neighbor of $\mathcal T_{q^{i+k}}$ for all $k \in \left\{0,\ldots, j-1 \right\}$, then
\begin{equation} \label{eq:deltaLnodeRelaton}
\Delta L_Y(\mathcal T_{q^i}, \mathcal T_{q^{i+j}};\beta)=\sum_{z \in  \mathcal B_{ij}}\Delta \hat L_Y(z;\beta),
\end{equation}
where $\mathcal B_{ij} = \mathcal N_{\text{int}}(\mathcal T_{q^{i+j}}) \setminus \mathcal N_{\text{int}}(\mathcal T_{q^{i}})$. 
Moreover, we should note the connection between \eqref{eq:deltaLnodeRelaton} and \eqref{eq:seqCharL}.
Namely, it can be shown that 
\begin{equation} \label{eq:lagrangianWorldRoot}
    L_Y(\mathrm{Root} \left(\mathcal T_{\mathcal W} \right); \beta ) = 0,
\end{equation}
which follows from the non-negativity of the mutual information and the properties of the entropy.
Taking $\mathcal T_{q^0} = \mathrm{Root}\left(\mathcal T_{\mathcal W}\right)$ in \eqref{eq:seqCharL} and utilizing \eqref{eq:lagrangianWorldRoot}, we see that for any $\mathcal T_{q^m} \in \mathcal T^{\mathcal Q}$,
\begin{equation} \label{eq:lagrangianAsDeltaL}
    L_Y(\mathcal T_{q^m};\beta) =  \sum_{i=0}^{m-1} \Delta\L_Y(\mathcal T_{\q^i},\mathcal T_{\q^{i+1}};\beta).
\end{equation}
Then, since \eqref{eq:deltaLnodeRelaton} provides a relation for the right-hand side of \eqref{eq:lagrangianAsDeltaL} we have, for any $\mathcal T_{q} \in \mathcal T^{\mathcal Q}$,
\begin{equation} \label{eq:lagrangianNodeDeltaL}
    L_Y(\mathcal T_{q};\beta) = \sum_{z \in  \mathcal N_{\text{int}}\left( \mathcal T_{q}\right)} \Delta \hat L_Y(z;\beta),
\end{equation}
since $\mathcal N_{\text{int}}\left(\mathrm{Root}\left( \mathcal T_{\mathcal W}\right) \right) = \emptyset$, which follows from Definition \ref{def:nodeProp}.
Thus, we see from \eqref{eq:lagrangianNodeDeltaL} that the value of $L_{Y}\left(\mathcal T_q ;\beta\right)$ for any tree $\mathcal T_q \in \mathcal T^{\mathcal Q}$ and $\beta > 0$ is the sum of the node-wise $\Delta \hat L_Y(\cdot;\beta)$ function over the interior nodes of the tree $\mathcal T_q \in \mathcal T^{\mathcal Q}$.
With this in place, we now have the following two lemmas, which will be useful for proving the optimality of the Q-tree search algorithm.
\begin{lemma} \label{lem:qtree2}
	\normalfont{
	Let $t \in \mathcal N_{\text{int}}(\mathcal T_{\mathcal W})$.
	Then $\hat Q_Y(t;\beta) > 0$ if and only if there exists a tree $\mathcal T_{q} \in \mathcal T^{\mathcal Q}$ such that $\sum_{z \in \mathcal N_{\text{int}}\left(\mathcal T_{q(t)}\right)} \Delta \hat L_Y(z;\beta)>0$.
	Furthermore, if $\hat Q_Y(t;\beta) > 0$, then there exists a tree $\mathcal T_{q^*} \in \mathcal T^{\mathcal Q}$ such that $\sum_{z \in \mathcal N_{\text{int}}\left(\mathcal T_{q^*(t)}\right)} \Delta \hat L_Y(z;\beta) = \hat Q_Y(t;\beta)$, and for all other trees $\mathcal T_{q'} \in \mathcal T^{\mathcal Q}$ with $t \in \mathcal N(\mathcal T_{q'})$ and $\mathcal T_{q'(t)} \neq \mathcal T_{q^*(t)}$ it holds that $\sum_{z \in \mathcal N_{\text{int}}\left(\mathcal T_{q'(t)}\right)} \Delta \hat L_Y(z;\beta) \leq \hat Q_Y(t;\beta)$.
	}
\end{lemma}

The following result implies that if a node with positive $\hat Q_Y(\cdot;\beta)$ is not expanded, then the resulting tree is sub-optimal with respect to \eqref{eq:IB_tree_optim_prob_2}.
\begin{lemma} \label{lem:Qsearch2}
	\normalfont{
	Let $\mathcal T_{q^*} \in \mathcal T^{\mathcal Q}$ be the solution returned by the Q-tree search algorithm and let $\mathcal T_{q'} \in \mathcal T^{\mathcal Q}$ be such that $\mathcal T_{q'} \subset \mathcal T_{q^*}$. 
	Then
	\begin{equation*}
	L_Y\left(\mathcal T_{q'};\beta \right) < L_Y\left(\mathcal T_{q^*};\beta \right).
	\end{equation*}
	}
\end{lemma}

Thus, Lemma \ref{lem:qtree2} establishes that a node with $\hat Q_Y(\cdot;\beta) > 0$ should be expanded, whereas Lemma~\ref{lem:Qsearch2} states that if the
nodes with $\hat Q(\cdot;\beta) > 0$ are not expanded then the resulting tree is sub-optimal with respect to $L_Y(\cdot;\beta)$.
The next theorem formally establishes the optimality of solutions found by the Q-tree search algorithm.

\begin{theorem} \label{thm:Qsearch3}
	\normalfont{
	Let $\mathcal T_{\tilde q} \in \mathcal T^{\mathcal Q}$ to be a minimal tree that is also optimal with respect to the cost $L_Y(\cdot;\beta)$. 
	Assume, 
	without loss of generality\footnote{The fully abstracted tree with single node $\mathrm{Root}(\mathcal{T}_{\mathcal{W}})$ is a subtree of any quadtree}, that the Q-tree search algorithm is initialized at the tree $\mathcal T_{q^0} \in \mathcal T^{\mathcal Q}$, where $\mathcal T_{q^0} \subseteq \mathcal T_{\tilde q}$ and 
	let $\mathcal T_{q^*} \in \mathcal T^{\mathcal Q}$ be the solution returned by the Q-tree search algorithm. 
	Then $\mathcal T_{q^*}=\mathcal T_{\tilde q}$.
	}
\end{theorem}

Theorem~\ref{thm:Qsearch3} establishes that the Q-tree search will find the globally optimal tree with respect to the cost $L_Y(\cdot;\beta)$, provided the algorithm is initiated at a tree $\mathcal T_{q^0} \in \mathcal T^{\mathcal Q}$ such that $\mathcal T_{q^0} \subseteq \mathcal T_{\tilde q}$.
Therefore, by selecting $\mathcal T_{q^0} = \mathrm{Root}(\mathcal T_{\mathcal W})$ we can guarantee that the Q-tree search algorithm will find the globally optimal solution.
Having established these results, we now discuss some details of our framework before demonstrating the approach with a numerical example. 

\subsection{Influence of $p(x,y)$} \label{sec:influenceOfPxy}

A tacit assumption regarding the probability distribution $p(x,y)$ has been made in the development of this framework.
Namely, provided that $p(x) > 0$, we can write the distribution $p(x,y)$ as $p(x,y) = p(y|x) p(x)$.
This poses no technical concern in the case that $p(x) > 0$ for all $x \in \Omega_X$.
In contrast, when $p(x) \ngtr 0$ for all $x \in \Omega_X$, it may occur that an aggregate node and all of its children nodes have no probability mass, which arises if $p(x) = 0$ for all $x \in \Omega_X$ that belong to the aggregate node $t \in \Omega_{T}$.
In this case, we have from \eqref{eq:agg_info_bottleneck_p_t} that $p(t) = 0$, but it is not clear that \eqref{eq:delta_L_agglomerative_IB} is well-defined.
Additionally, the need to investigate this scenario is clear from Definition~\ref{def:deltaLhat} and the subsequent discussion, as it illustrates the connection between the change in the objective function value when moving from tree $\mathcal T_{q^i} \in \mathcal T^{\mathcal Q}$ to tree $\mathcal T_{q^{i+1}} \in \mathcal T^{\mathcal Q}$ to the node-specific quantities.
Thus, in order to apply the Greedy or Q-tree search algorithms for general $p(x)$, we must establish that \eqref{eq:delta_L_agglomerative_IB} is well defined in these cases.
This leads us to the following proposition.

\begin{proposition} \label{prop:limitDeltaL}
	\normalfont{
	Let  $t \in \mathcal N_{\text{int}}(\mathcal T_{\mathcal W})$ and assume $p(x) = {\varepsilon}{/N}$ for all $x \in \mathcal N_{\text{leaf}}(\mathcal T_{\mathcal W(t)})$ with $N = \lvert \mathcal N_{\text{leaf}}(\mathcal T_{\mathcal W(t)}) \rvert$ for some $\varepsilon > 0$.
	Then $\lim\limits_{\varepsilon \to 0^{+}} \Delta \hat L(t;\beta) = 0$ for all $\beta > 0$.
	}
\end{proposition}
\vspace{0.1cm}

The utility of Proposition~\ref{prop:limitDeltaL} is that it allows for the direct application of both the Greedy and Q-tree search algorithms for any $p(x)$ without modification to the respective algorithms.
This allows us not only to form abstractions as a function of $\beta > 0$, but lets us also dictate \emph{where} information is important by changing $p(x)$.
To see why $p(x)$ allows us to dictate where information is important, let the joint distribution $p(x,y)$ be defined by $p(y|x)$ and $p(x)$ as $p(x,y) = p(y|x)p(x)$ and consider
\begin{equation} \label{eq:infPxyDecoder1}
p(y|t) = \frac{1}{p(t)} \sum_{x \in \mathcal N_{\text{leaf}}(\mathcal T_{\mathcal W(t)})}p(y|x) p(x).
\end{equation}
From \eqref{eq:infPxyDecoder1} we see that nodes $x \in \mathcal N_{\text{leaf}}(\mathcal T_{\mathcal W(t)})$ that are aggregated to $t \in \mathcal N_{\text{int}}(\mathcal T_{\mathcal W})$ and have $p(x) = 0$ do not contribute to the conditional distribution $p(y|t)$, and thus have lower importance to the optimization problem as these nodes convey no information regarding $Y$.
Thus, abstract nodes $t \in \Omega_T$ for which the underlying $x\in  \mathcal N_{\text{leaf}}(\mathcal T_{\mathcal W(t)})$ have high $p(y|x)$ and $p(x)$ will have the greatest information context regarding $Y$, since these conditions will increase the value of $p(y|t)$.
Furthermore, we see from \eqref{eq:infPxyDecoder1} that, when $p(x)$ is uniform, the algorithm does not discriminate as to where the information in the environment is located, as each value of $p(y|x)$ for $x \in \mathcal N_{\text{leaf}}(\mathcal T_{\mathcal W(t)})$ is given equal weight when computing $p(y|t)$. 
Consequently, as $\beta \to \infty$ the algorithms become concerned with retaining all the relevant information in the environment, regardless of where this information is located.
This is shown in the numerical example we discuss next.


\section{Numerical Example} \label{sec:Results_and_Discuss}

\begin{figure}[!hbt]
	\centering
	\includegraphics[scale=0.25]{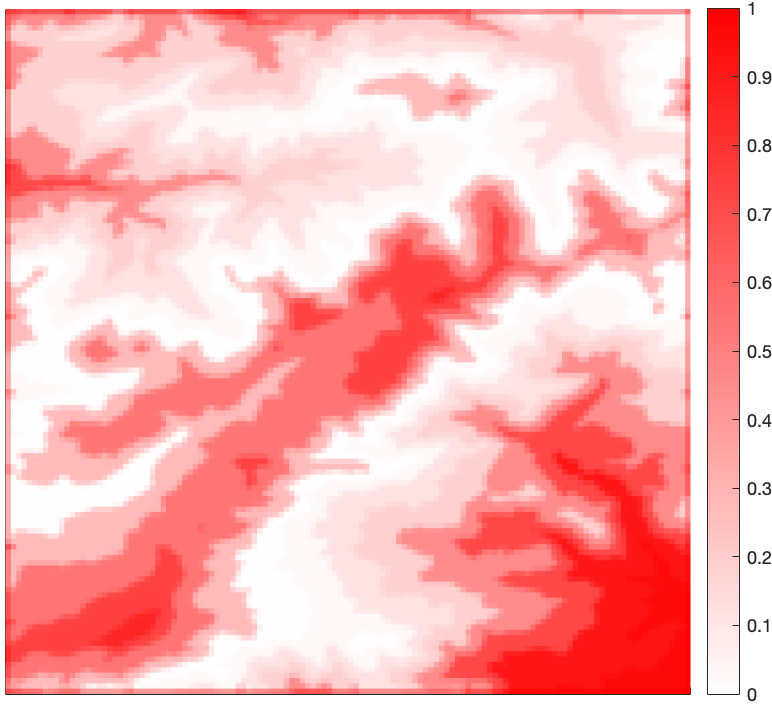}
	\caption{128$\times$128 original map of environment.  Shading of red indicates the probability that a cell is occupied.}
	\label{fig:original_environment}
\end{figure}

In this section, we present a numerical example to demonstrate the emergence of abstractions in a grid-world setting.
To this end, consider the environment shown in Figure \ref{fig:original_environment} having dimension 128$\times$128.
We view this map as representing an environment where the intensity of the color indicates the probability that a given cell is occupied.
In this view, the map in Figure \ref{fig:original_environment} can be thought of as an occupancy grid (OG) where the original space, $X$, is considered to be the elementary cells shown in the figure.
We wish to compress $X$ to an abstract representation $T$ (a quadtree), while preserving as much information regarding cell occupancy as possible.
Thus, we take the relevant random variable, $Y$, as the probability of occupancy and study this problem while varying $\beta > 0$.
Therefore, $\Omega_Y = \left\{0,1\right\}$ where $y = 0$ corresponds to free space and $y = 1$ occupied space.
It is assumed that $p(x)$ is provided and $\p(y|x)$ is given by the occupancy grid, where $\p(x,y) = \p(y|x) \p(x)$ .
\subsection{Region-Agnostic Abstraction} \label{sec:uniformPxSection}

\begin{figure}[b]
	\centering
	\includegraphics[scale=0.12]{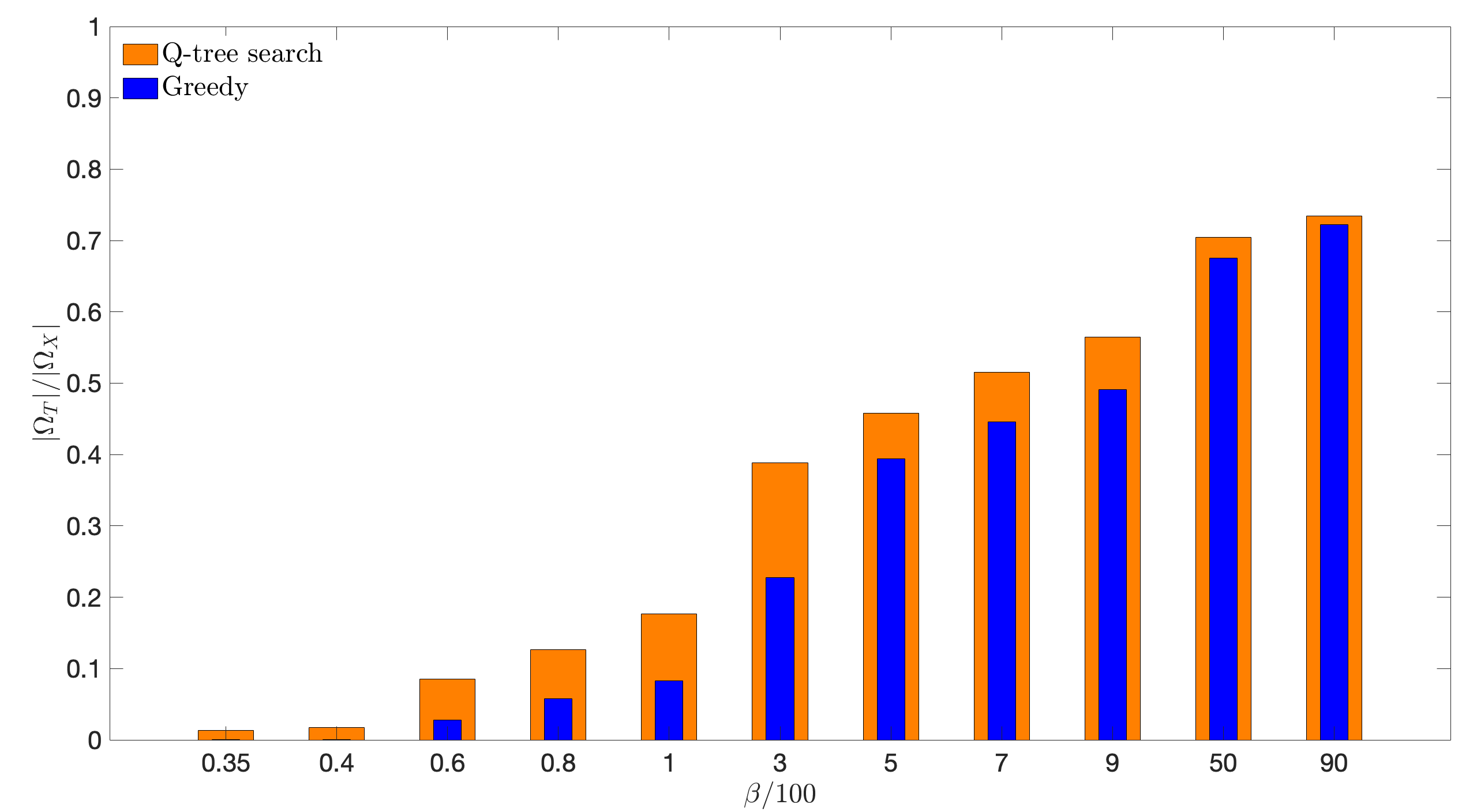}
	\caption{$\nicefrac{\lvert \Omega_T \rvert}{\lvert \Omega_X \rvert}$ vs. $\nicefrac{\beta}{100}$ for the Greedy and Q-tree search algorithms, $\lvert \Omega_X \rvert = 16384$.}
	\label{fig:relative_card_bar}
\end{figure}

In this section, we assume that $\p(x)$ is uniform.
By changing $\beta$  we obtain a family of solutions, with the leaf node cardinality of the resulting tree returned by the respective algorithm shown in Figure \ref{fig:relative_card_bar}.
\begin{figure}[t]
	\centering
	\includegraphics[scale=0.12]{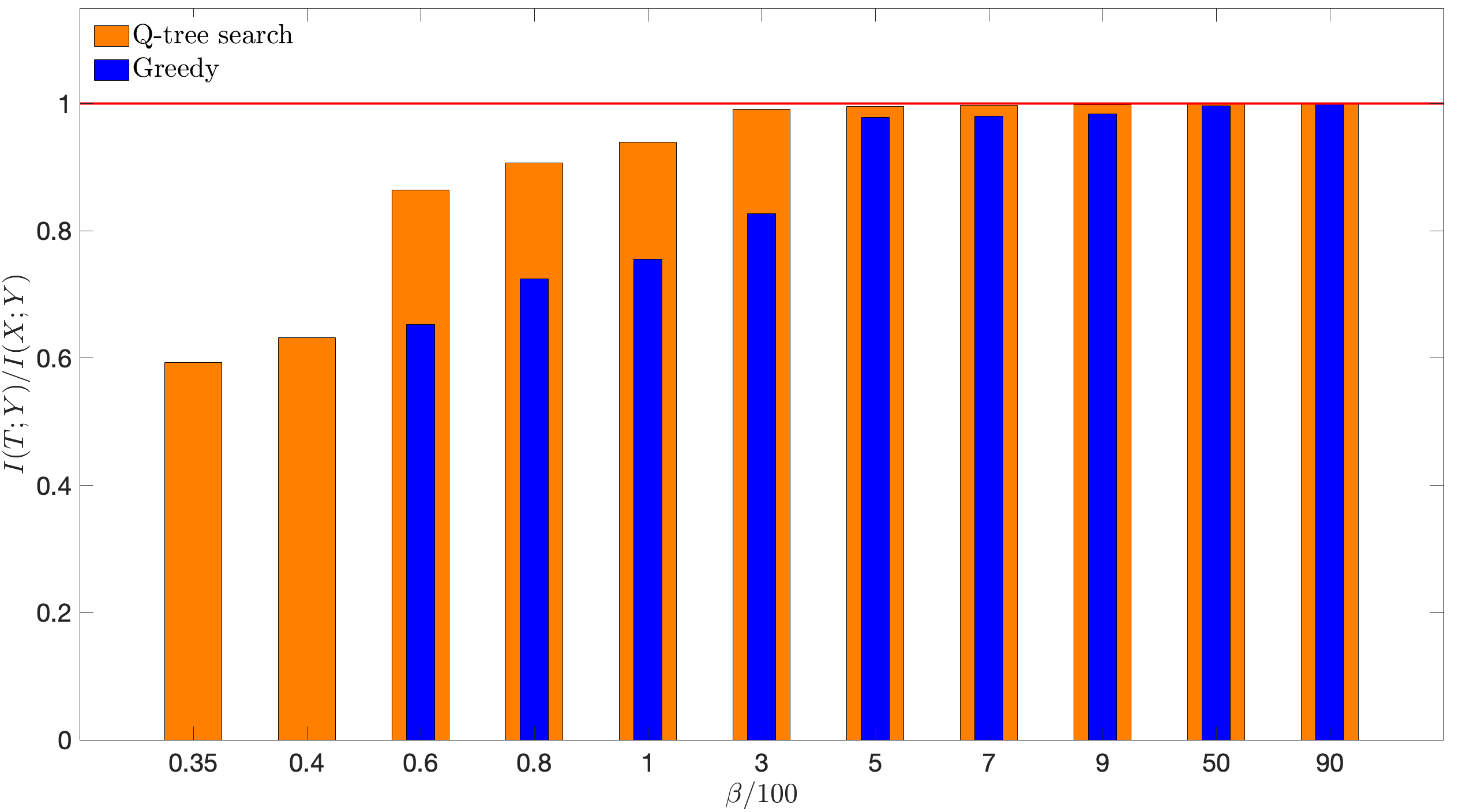}
	\caption{$\nicefrac{\I(T;Y)}{\I(Y;X)}$ vs. $\nicefrac{\beta}{100}$ for the Greedy and Q-tree search algorithms.}
	\label{fig:I_TY_greedy_and_Q}
\end{figure}
\begin{figure}[htb]
	\centering
	\includegraphics[scale=0.12]{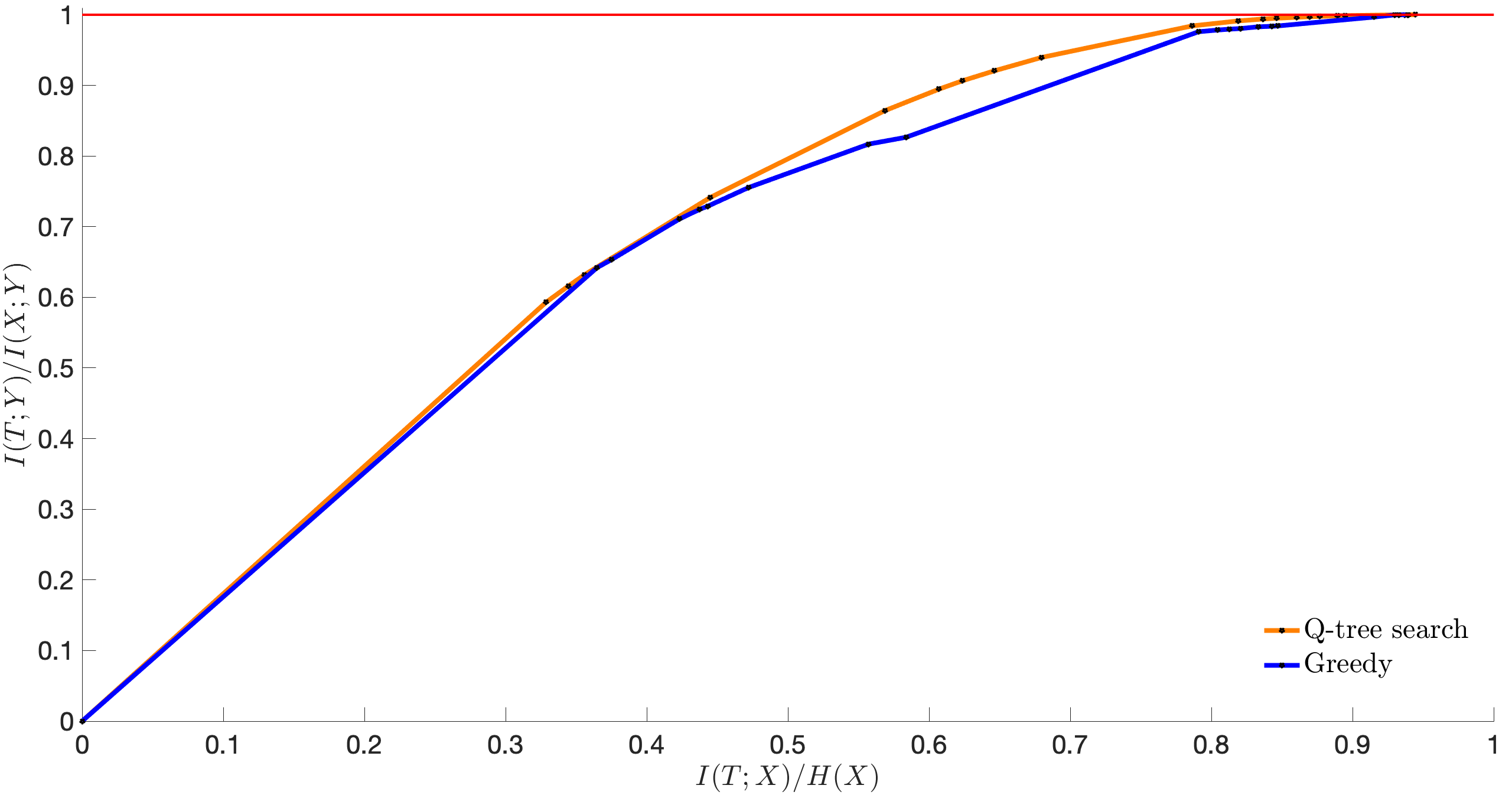}
	\caption{Information plane for Greedy and Q-tree search algorithms.}
	\label{fig:Info_plane}
\end{figure}
As seen in Figure \ref{fig:relative_card_bar}, the number of leaf nodes of the trees found by both algorithms is increasing with $\beta$.
Furthermore,  the Q-tree search and Greedy leaf node cardinalities converge as $\beta$ tends toward infinity, as expected.
Additionally, as seen in Figure \ref{fig:I_TY_greedy_and_Q}, the information contained in the compressed representation $T$ regarding the relevant variable $Y$, given by $\I(T;Y)$, approaches the information that the original space $X$ contains about $Y$, quantified by $\I(X;Y)$.
Note also that $\I(T;Y) \leq \I(X;Y)$, which follows from the Markov chain $Y \to X \to T$ and the data processing inequality.
This encodes the fact that the information contained about the relevant variable $Y$ retained by the abstraction $T$ cannot exceed that given by the original space $X$.
Furthermore, from Figure \ref{fig:I_TY_greedy_and_Q}, we notice that the Q-tree search algorithm finds solutions that are more informative regarding the relevant variable $Y$ than the Greedy algorithm, indicating that the Greedy algorithm terminates prematurely, and that further improvement is possible for the given $\beta > 0$.
We also see that the solutions of the Greedy algorithm and of the Q-tree search converge as $\beta$ approaches infinity.

\begin{figure}[htb]
	\centering
	\begin{minipage}{.49\linewidth}
		\centering
		\includegraphics[scale=0.19]{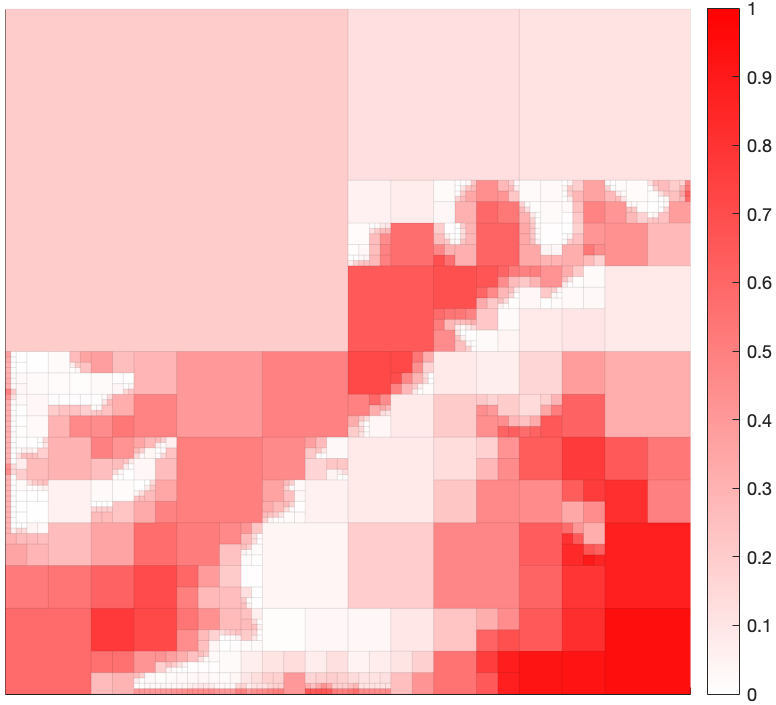}
		\caption{$\beta = 50$ representation.}
		\label{fig:abs_vis_beta_50}
	\end{minipage}
	\hfill
	\begin{minipage}{.49\linewidth}
		\centering
		\includegraphics[scale=0.19]{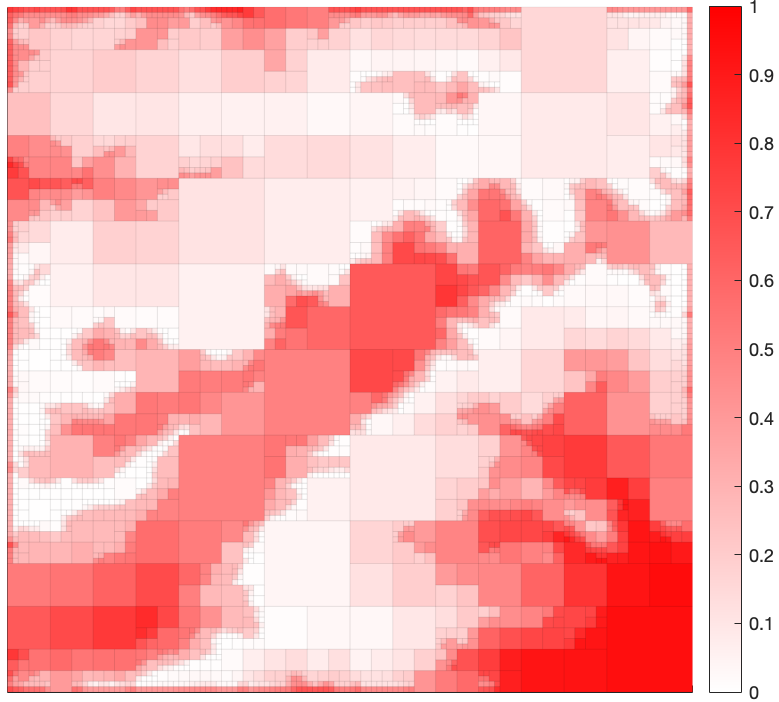}
		\caption{$\beta = 100$ representation.}
		\label{fig:abs_vis_beta_100}
	\end{minipage}
\end{figure}
\begin{figure}[htb]
	\centering
	\begin{minipage}{.49\linewidth}
		\centering
		\includegraphics[scale=0.19]{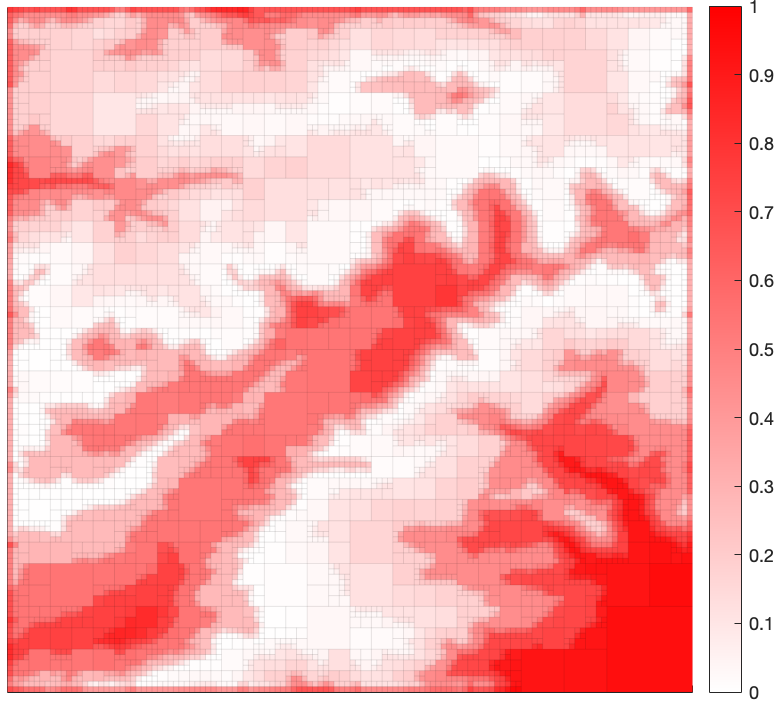}
		\caption{$\beta = 400$ representation.}
		\label{fig:abs_vis_beta_400}
	\end{minipage}
	\hfill
	\begin{minipage}{.49\linewidth}
		\centering
		\includegraphics[scale=0.19]{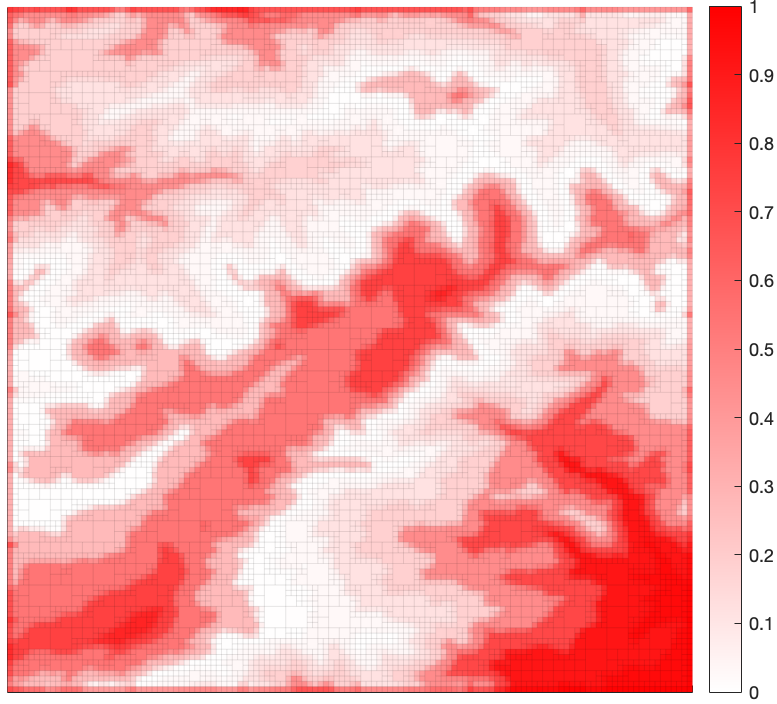}
		\caption{$\beta = 15000$ representation.}
		\label{fig:abs_vis_beta_15000}
	\end{minipage}
\end{figure}

Shown in Figure \ref{fig:Info_plane} is the information plane, where the normalized $\I(T;Y)$ is plotted versus the normalized $\I(T;X)$.
In this way, the information plane displays the amount of relevant information retained in a solution vs. the level of compression of $X$.
In viewing this figure, recall that Theorem~\ref{thm:Qsearch3} establishes the global optimality of solutions obtained by Q-tree search, and hence no solution above the Q-tree search line is possible in the space $\mathcal T^{\mathcal Q}$, since this would imply that solutions (trees) encoding more information about $Y$, and for the same level of compression, exist in $\mathcal T^{\mathcal Q}$.

With this in mind, Figure \ref{fig:Info_plane} also corroborates that the Greedy algorithm generally finds solutions that are sub-optimal with respect to $L_Y(\cdot;\beta)$, since trees found by the Greedy algorithm retain less information about $Y$ for the same level of compression as the information-plane curve of Greedy lies below that of Q-tree search. 
Moving along the curve is done by varying $\beta$, with increasing $\beta$ moving the solution to the right in this plane, towards more informative, higher cardinality solutions.
We can see from Figure~\ref{fig:Info_plane} the advantage of utilizing the Q-tree search algorithm, as the Greedy approach arrives at solutions that are sub-optimal compared  to those found by the Q-tree search algorithm.
A sample of environment depictions for various values of $\beta$ obtained from the Q-tree search algorithm are shown in 
Figures~\ref{fig:abs_vis_beta_50}-\ref{fig:abs_vis_beta_15000}.
As seen in these figures, the solution returned by the Q-tree search algorithm approaches that of the original space as $\beta \to \infty$, with a spectrum of solutions obtained as $\beta$ is varied.  
These figures show that areas containing high information content, as specified by $Y$, are refined first while leaving the regions with less information content to be refined at a higher $\beta$.

We see that $\beta$ resembles a sort of a ``gain" that can be increased, resulting in progressively more informative solutions of higher cardinality.
Thus, once the map is given, changing \emph{only} the value of $\beta$ gives rise to a variety of solutions of varying resolution.
That is, our framework finds the optimal tree $\mathcal T_{q^*}$ with respect to $L_Y(\cdot;\beta)$ without the need to specify pre-defined pruning rules or a host of parameters that define the granularity of the abstraction a priori.
Interestingly, $\beta$ plays a similar role in this work as in \cite{Larsson2017,Tishby2010,Genewein2015}.
Namely, as $\beta \to 0$, highly compressed representations of the space are obtained whereas for large values of $\beta$, 
we asymptotically approach the original map.
Thus, we can view $\beta$ as a ``rationality parameter," analogous to \cite{Larsson2017,Tishby2010,Genewein2015}, where agents with low $\beta$ are considered to be more resource limited, thus utilizing simpler, lower cardinality representations of the environment.

\subsection{Region-Specific Abstraction}
\begin{figure}[h!]
	\centering
	\includegraphics[scale=0.25]{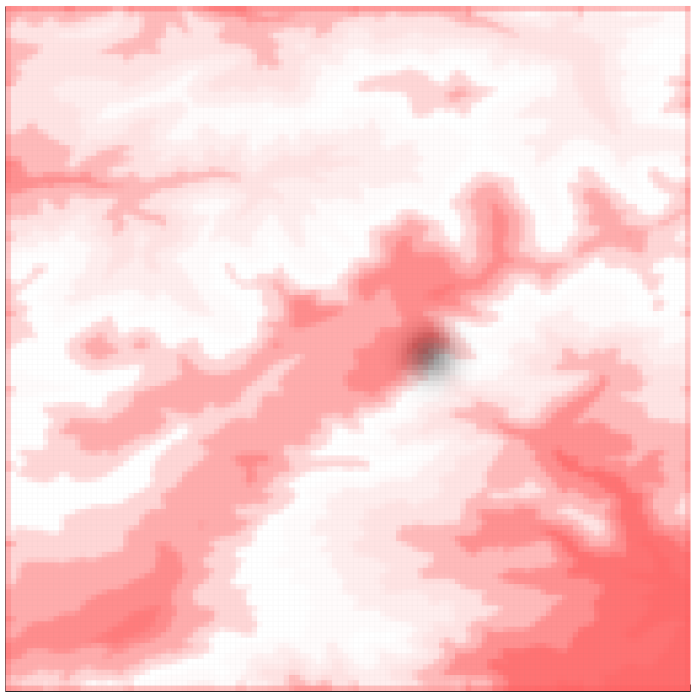}
	\caption{128$\times$128 original map of environment with overlayed $p(x)$.  Shading of red indicates the probability that a cell is occupied by an obstacle, whereas the shade of black indicates cell probability mass under $p(x)$.} 
	\label{fig:nonUniformPx}
\end{figure}
In the previous section, we discussed how the Greedy and Q-tree search algorithms can be used to obtain abstractions as a function of $\beta > 0$ under the assumption that the distribution $p(x)$ is uniform.
We now relax this assumption and discuss the ability to obtain region-specific abstractions in the environment through a non-uniform $p(x)$, without modification to the underlying framework or algorithms as discussed in Section \ref{sec:influenceOfPxy}.
\begin{figure}[!htb]
	\centering
	\includegraphics[scale=0.12]{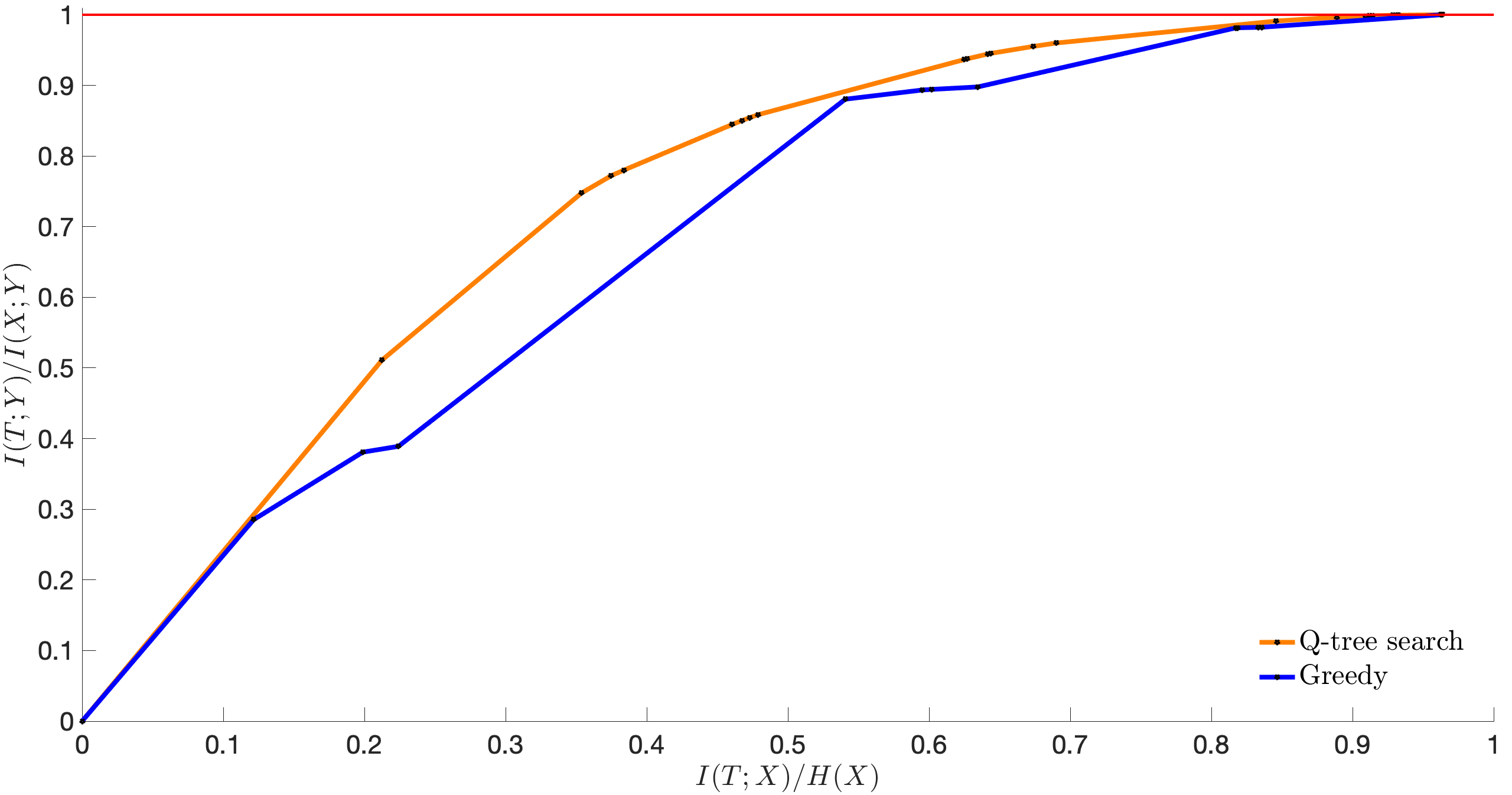}
	\caption{Information plane for Greedy and Q-tree search algorithms, non-uniform $p(x)$.}
	\label{fig:info_plane_non_uniPx}
\end{figure}
\begin{figure}[h!]
	\centering
	\includegraphics[scale=0.125]{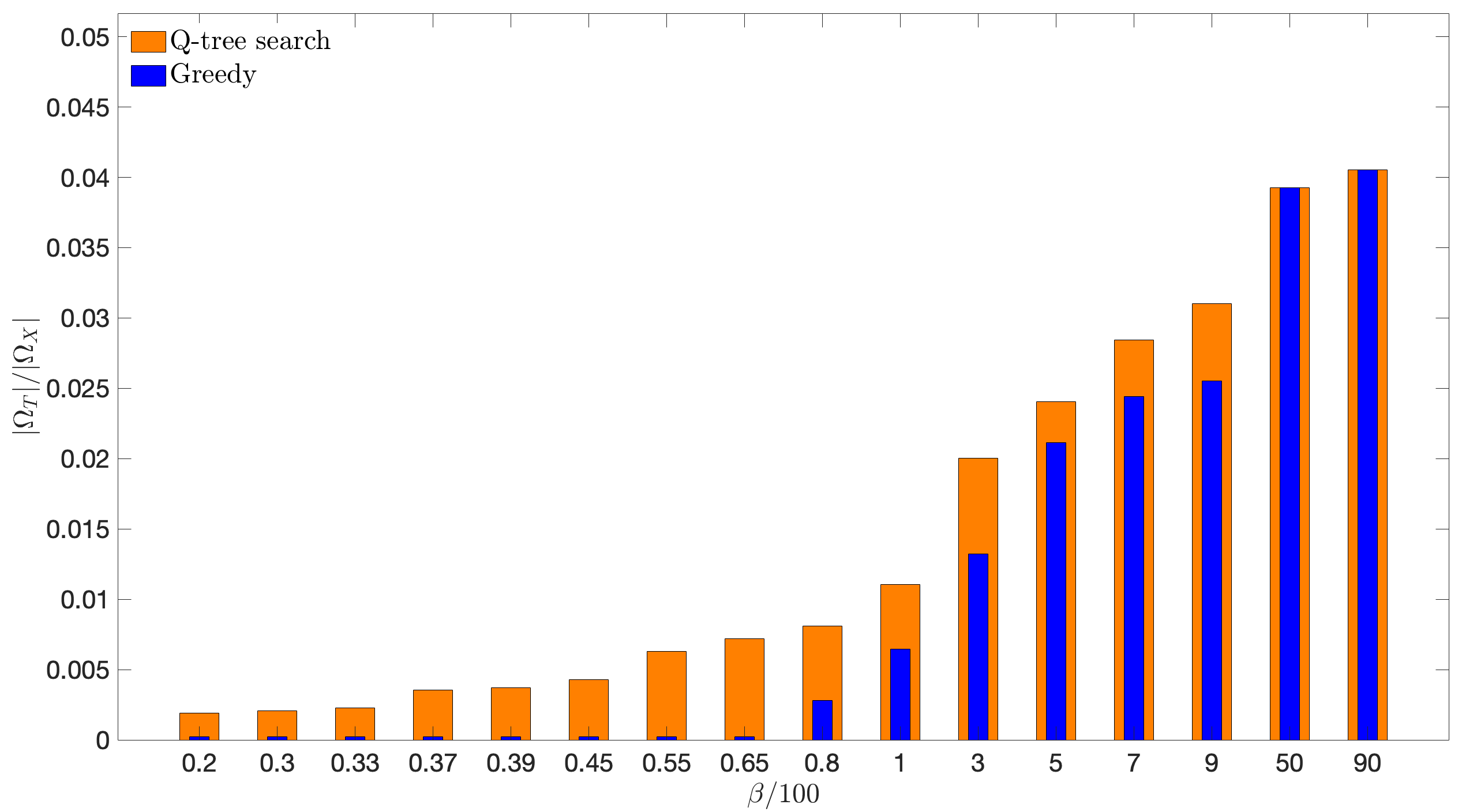}
	\caption{$\nicefrac{\lvert \Omega_T \rvert}{\lvert \Omega_X \rvert}$ vs. $\nicefrac{\beta}{100}$ for Greedy and Q-tree search algorithms, non-uniform $p(x)$. Note y-axis scaling, $\lvert \Omega_X \rvert = 16384$.}
	\label{fig:cardinality_non_uniPx}
\end{figure}
We utilize the same environment as in Figure \ref{fig:original_environment}, but with a non-uniform distribution $p(x)$, as shown in Figure~\ref{fig:nonUniformPx}.
In this example, we take $p(x)$ to be a two-dimensional Gaussian distribution with mean $\mu = \left[80, 63\right]^\tp$ and covariance matrix $\Sigma = 10 I_{2 \times 2}$.
For comparison, we obtain solutions from both the Greedy and Q-tree search algorithms for a range of $\beta$-values.
\begin{figure}[!htb]
	\centering
	\begin{minipage}{.49\linewidth}
		\centering
		\includegraphics[scale=0.19]{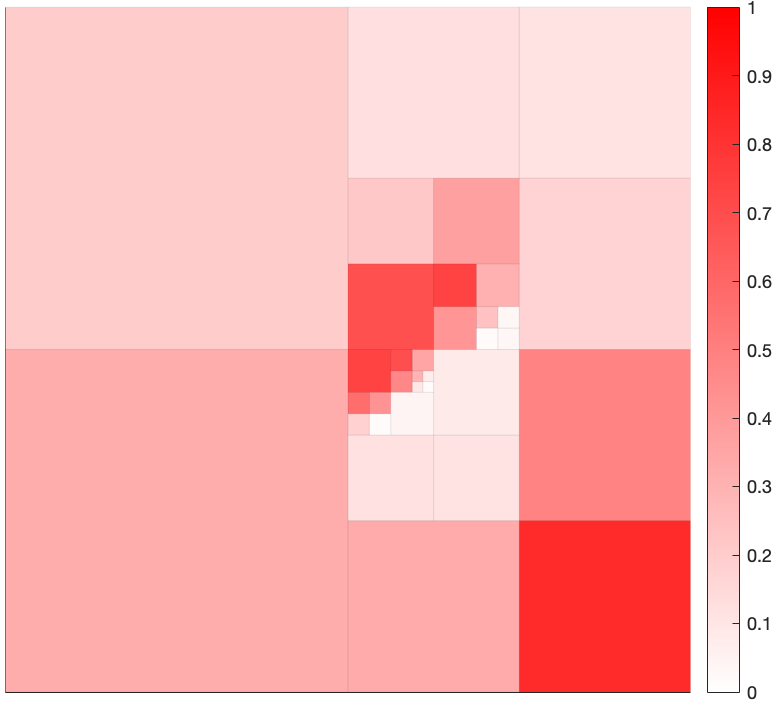}
		\caption{$\beta = 25$ representation.}
		\vspace*{6 pt}
		\label{fig:abs_vis_beta_25_nonUniPx}
	\end{minipage}
	\hfill
	\begin{minipage}{.49\linewidth}
		\centering
		\includegraphics[scale=0.19]{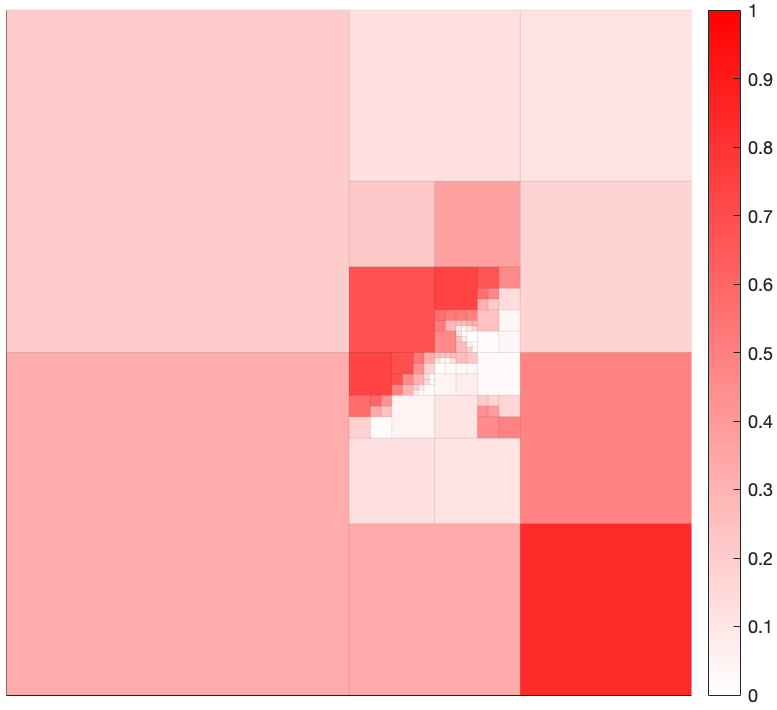}
		\caption{$\beta = 55$ representation.} 
		\vspace*{6 pt}
		\label{fig:abs_vis_beta_55_nonUniPx}
	\end{minipage}
	~
	\centering
	\begin{minipage}{.49\linewidth}
		\centering
		\includegraphics[scale=0.19]{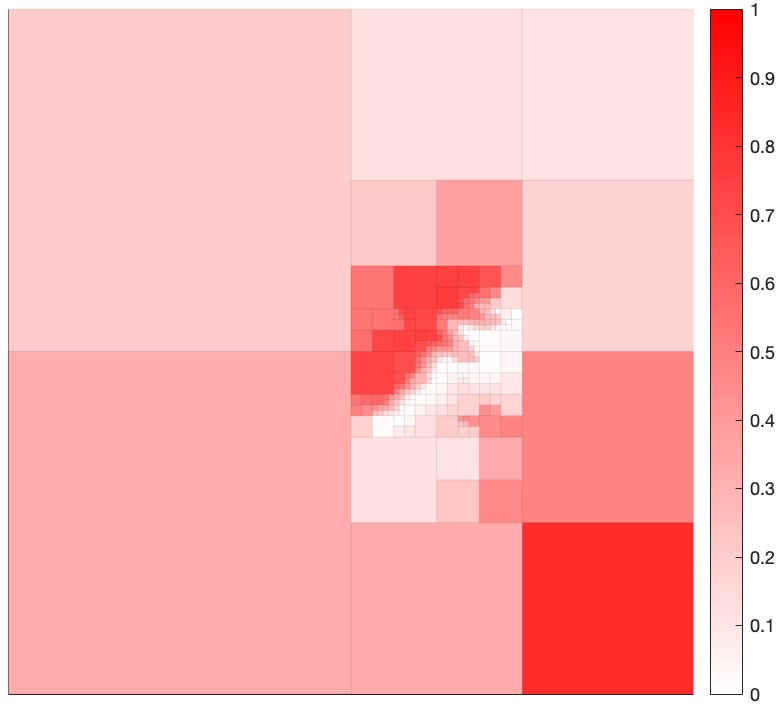}
		\caption{$\beta = 200$ representation.}
		\label{fig:abs_vis_beta_200_nonUniPx}
	\end{minipage}
	\hfill
	\begin{minipage}{.49\linewidth}
		\centering
		\includegraphics[scale=0.19]{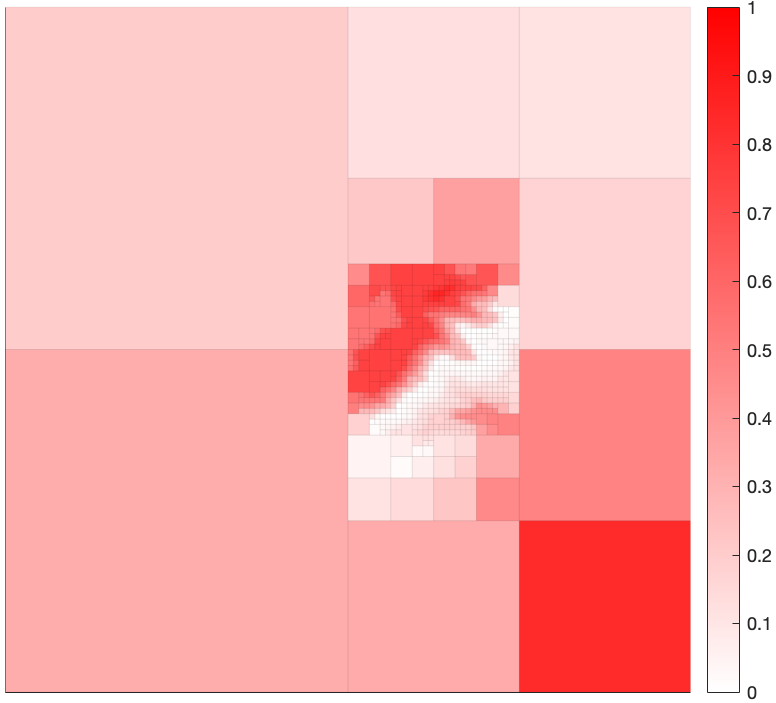}
		\caption{$\beta = 15000$ representation.}
		\label{fig:abs_vis_beta_15000_nonUniPx}
	\end{minipage}
\end{figure}
The information plane is shown in Figure \ref{fig:info_plane_non_uniPx} with the cardinality of the resulting tree in Figure \ref{fig:cardinality_non_uniPx}.
We see from Figure \ref{fig:info_plane_non_uniPx} that the Greedy algorithm finds solutions that are sub-optimal with respect to Q-tree search, since for a given level of compression ($I(T;X)$), the Greedy algorithm finds solutions that are less informative about $Y$.
Figure~\ref{fig:cardinality_non_uniPx} shows that the Q-tree search algorithm finds solutions that are of higher leaf-node cardinality than those found by Greedy, but that the solutions returned by Q-tree search contain more relevant information.
Figures \ref{fig:relative_card_bar} and \ref{fig:cardinality_non_uniPx} differ due to the difference in $p(x)$ in the sense 
that regions with $p(x) = 0$ do not contain any information regarding $Y$, as seen by \eqref{eq:infPxyDecoder1} and the subsequent discussion.
Finally, visualizations of the resulting solutions obtained from the Q-tree search algorithm are provided in 
Figures~\ref{fig:abs_vis_beta_25_nonUniPx}-\ref{fig:abs_vis_beta_15000_nonUniPx}.
These figures corroborate the previous observations, where we can clearly see that the algorithm refines only regions for which $p(x) > 0$.
Furthermore, the refinement is progressive and of increasing resolution as $\beta \to \infty$.


\section{Conclusions} \label{sec:Conclusions}

In this paper, we have developed a novel framework for the emergence of abstractions that are not provided to the agent a priori but instead arise as a result of the available 
agent computational resources.
We utilize concepts from information theory, such as the information bottleneck and agglomerative information bottleneck methods to formulate a new optimization problem over the space of trees.
The structural properties of the framework were discussed with applications to bounded rationality and information-limited agents.
Finally, we propose and analyze two algorithms, which were implemented in order to obtain solutions for a two-dimensional environment.

The importance of this work lies in the development of a framework that allows for the emergence of abstractions in a principled manner.
The proposed algorithms demonstrate the utility of the approach, requiring only the specification of a relevant variable that contains the information we wish to retain in the resulting compressed representation.
The framework then searches for trees that not only compress the 
original space, but maximally preserve the information regarding the relevant variable.
The results can be utilized in decision-making problems to systematically 
compress the given state representation or in path-planning algorithms to develop
 reduced complexity representations of the original planning space.

\section*{Acknowledgement}	

Support for this work has been provided by ONR awards N00014-18-1-2375 and N00014-18-1-2828
and by ARL under DCIST CRA W911NF-17-2-018.


\bibliographystyle{IEEEtran}
\bibliography{ms}


\section*{Appendix}

\subsection{Proof of Theorem \ref{thm:Q_search_and_greedy_soln}}
	Note that 
	\begin{align} \label{eq:inequality_delta_L}
	\Delta \L_Y(&\mathcal T_{\q^i},\mathcal T_{\q^{i+1}};\beta) \leq  \nonumber \\
	&\Delta \L_Y(\mathcal T_{\q^i},\mathcal T_{\q^{i+1}};\beta) + \sum_{\tau=1}^n \Q_Y(\mathcal T_{\q^{i+1}},\mathcal T_{\q_{\tau}^{i+2}};\beta),
	\end{align}
	since $\Q_Y(\mathcal T_{\q^{i+1}},\mathcal T_{\q_\tau^{i+2}};\beta) \geq 0$. 
	In the Greedy algorithm, a node is expanded, adding $\left\{ t'_{1},\ldots,t'_{n} \right\}$ to $\mathcal{N}(\mathcal T_{\q^i})$ to obtain $\mathcal N(\mathcal T_{\q^{i+1}})$, if $\Delta \L_Y(\mathcal T_{\q^i},\mathcal T_{\q^{i+1}};\beta) > 0$.
	If $\Delta \L_Y(\mathcal T_{\q^i},\mathcal T_{\q^{i+1}};\beta) > 0$ then by \eqref{eq:inequality_delta_L} and \eqref{eq:Q_Y_future} it follows that
	\begin{align*}
	0 < \Delta \L_Y(\mathcal T_{\q^i},\mathcal T_{\q^{i+1}};\beta) + \sum_{\tau=1}^n \Q_Y(\mathcal T_{\q^{i+1}},\mathcal T_{\q_\tau^{i+2}};\beta &) \\
	= \Q_Y(\mathcal T_{\q^{i}},\mathcal T_{\q^{i+1}};\beta&) ,
	\end{align*}
	and therefore $\Q_Y(\mathcal T_{\q^{i}},\mathcal T_{\q^{i+1}};\beta) > 0$.
	Hence nodes expanded by the Greedy algorithm will also be expanded by Q-tree search.
    Since the two algorithms are initialized at a common $\mathcal T_{q^0} \in \mathcal T^{\mathcal Q}$, it follows that $\mathcal T_{q_{\mathrm{G}}^*} \subseteq \mathcal T_{q_{\mathrm{Q}}^*}$.

\subsection{Proof of Lemma \ref{lem:qtree2}}
	The proof is given by induction. 
	We first establish necessity and sufficiency for some $t\in  \mathcal N_{\ell-1}(\mathcal T_{\mathcal W})$, where $\ell > 0$ is the maximum 
	depth of $\mathcal T_{\mathcal W}$.\\
	($\Rightarrow$)	Assume $\hat Q_Y(t;\beta) > 0$ for some $t \in \mathcal N_{\ell-1}(\mathcal T_{\mathcal W})$. 
	We thus have
	\begin{equation*}
	0 < \hat Q_Y(t;\beta) = \max\big\{\Delta \hat L_Y(t;\beta) + \sum_{t'\in \mathcal C (t)}\hat Q_Y(t';\beta);~0 \big\}.
	\end{equation*}
	Hence,
	\begin{equation*}
	\Delta \hat L_Y(t;\beta) + \sum_{t'\in \mathcal C(t)}\hat Q_Y(t';\beta) > 0.
	\end{equation*}
	Since $t \in \mathcal N_{\ell -1}(\mathcal T_{\mathcal W})$ it follows that $t' \in \mathcal C(t) \subset \mathcal N_{\text{leaf}}(\mathcal T_{\mathcal W})$ and thus $\hat Q_Y(t';\beta) = 0$, which implies that $\Delta \hat L_Y(t;\beta) = \hat Q_Y(t;\beta) > 0$. 
	Now consider the tree $\mathcal T_{q} \in \mathcal T^{\mathcal Q}$ such that $\mathcal N_{\text{leaf}}(\mathcal T_{q(t)}) = \mathcal C(t)$.
	Then, for the subtree $\mathcal T_{q(t)} \subseteq \mathcal T_{\mathcal W}$ 
	\begin{equation*}
	\sum_{z \in \mathcal N_{\text{int}}(\mathcal T_{q(t)})}\Delta \hat L_Y(z;\beta) = \Delta \hat L_Y(t;\beta) > 0.
	\end{equation*}
	($\Leftarrow$) Assume there exists a tree $\mathcal T_{q} \in \mathcal T^{\mathcal Q}$ such that
	\begin{equation*}
	\sum_{z \in \mathcal{N}_{\text{int}}(\mathcal T_{q(t)})}\Delta \hat L_Y(z;\beta) > 0.
	\end{equation*}
	Note that, since $t \in \mathcal N_{\ell -1}(\mathcal T_{\mathcal W})$ then $\mathcal N (\mathcal T_{q(t)}) = \left\{ t \right\} \cup \mathcal C(t)$, with $\mathcal N_{\text{int}}(\mathcal T_{q(t)}) = \left\{ t\right\}$ and $\mathcal N_{\text{leaf}}(\mathcal T_{q(t)}) = \mathcal C(t)\subset \mathcal{N}_{\text{leaf}}(\mathcal T_{\mathcal W})$. 
	Therefore,
	\begin{equation*}
	0< \sum_{z \in \mathcal N_{\text{int}}(\mathcal T_{q(t)})}\Delta \hat L_Y(z;\beta) = \Delta \hat L_Y(t;\beta),
	\end{equation*}
	and 
	\begin{align*}
	\hat Q_Y(t;\beta) &= \max \big\{\Delta \hat L_Y(t;\beta) + \sum_{t'\in \mathcal C(t)} \hat Q_Y(t';\beta),~0 \big\}, \\
	&= \max \big\{\Delta \hat L_Y(t;\beta) ,~0 \big\}, \\
	&= \Delta \hat L_Y(t;\beta) > 0.
	\end{align*}
	Furthermore, for the tree $\mathcal{T}_{q}$ we have $\sum_{z \in \mathcal N_{\text{int}}(\mathcal T_{q(t)})}\Delta \hat L_Y(z;\beta)=\hat Q_Y(t;\beta)$ and
	 since $t\in \mathcal N_{\ell-1}(\mathcal T_{\mathcal W})$, for any other tree $\mathcal{T}_{\tilde q}$ such that $\mathcal{T}_{\tilde q(t)}\ne \mathcal{T}_{q(t)}$, it holds that $\mathcal N_{\text{int}}({\mathcal{T}_{\tilde q (t)}})=\emptyset$, which implies that $\sum_{z \in \mathcal N_{\text{int}}(\mathcal T_{\tilde q(t)})}\Delta \hat L_Y(z;\beta)=0 \le \hat Q_Y(t;\beta).$
	Thus, the lemma is true for all nodes $t \in \mathcal N_{\ell -1}(\mathcal T_{\mathcal W})$.
	\vspace{.2 cm}
	
	We now establish necessity and sufficiency for all $k \in \left\{1,\ldots, \ell - 1\right\}$.
	To this end, assume that for some $k \in \left\{1,\ldots, \ell - 1\right\}$ and  any $t' \in \mathcal N_k(\mathcal T_{\mathcal W})$,  $\hat Q_Y(t';\beta) >0$ if and only if there exists a tree $\mathcal{T}_q \in \mathcal{T}^{\mathcal{Q}}$ such that $\sum_{z\in \mathcal{N}_{\text{int}}(\mathcal{T}_{q(t')})}\Delta \hat L_Y(z;\beta)>0$.
	Furthermore, if $\hat Q_Y(t';\beta) >0$ then there exists a tree $\mathcal{T}_{q^*} \in \mathcal{T}^{\mathcal{Q}}$ such that $\sum_{z\in \mathcal{N}_{\text{int}}(\mathcal{T}_{q^*(t')})}\Delta \hat L_Y(z;\beta)=\hat Q_Y(t';\beta)$, and for all other trees $\mathcal{T}_{\tilde q} \in \mathcal{T}^{\mathcal{Q}}$ with $t' \in \mathcal{N}(\mathcal T_{\tilde q})$ and $\mathcal{T}_{\tilde q(t')} \ne  \mathcal{T}_{q^*(t')}$, $\sum_{z\in \mathcal{N}_{\text{int}}(\mathcal{T}_{\tilde q(t')})}\Delta \hat L_Y(z;\beta)\le \hat Q_Y(t';\beta)$. 
	Using this hypothesis, we prove that the lemma also holds for all $t \in \mathcal N_{k-1}(\mathcal T_{\mathcal W})$.

	\vspace{.2 cm}
	\noindent$\left( \Rightarrow \right)$
	Consider $t \in \mathcal N_{k-1}(\mathcal T_{\mathcal W})$ and assume that $\hat Q_Y(t;\beta) > 0$. 
	Define the set
	\begin{equation*}
	\mathcal S = \left\{t' \in \mathcal C (t) : \hat Q_Y(t';\beta) > 0 \right\} \subset \mathcal N_k(\mathcal T_{\mathcal W}).
	\end{equation*} 
    If $\mathcal{S}=\emptyset$ then from Definition \ref{def:node_wise_Q}, $0 < \hat Q_Y(t;\beta)=\max\{\Delta \hat L_Y(t;\beta),0\} $, and therefore $\hat Q_Y(t;\beta)=\Delta \hat L_Y(t;\beta) >0$.
    Now, consider any tree $\mathcal T_{q} \in \mathcal T^{\mathcal Q}$ such that $\mathcal T_{q(t)}$ has node set $\mathcal N (\mathcal T_{q(t)}) = \left\{ t \right\} \cup \mathcal C(t)$.
	Note that $\mathcal N_{\text{int}}(\mathcal T_{q(t)}) = \left\{ t\right\}$ and $\mathcal N_{\text{leaf}}(\mathcal T_{q(t)}) = \mathcal C(t)$.
	Thus, for the subtree $\mathcal T_{q(t)}$, 
	\begin{equation*}
	\sum_{z \in \mathcal N_\text{int}(\mathcal T_{q(t)})} \Delta \hat L_Y(z;\beta) = \Delta \hat L_Y(t;\beta) = \hat Q_Y(t;\beta).
	\end{equation*}
    Therefore $\hat Q_Y(t;\beta)>0$ implies that there exists a tree $\mathcal T_{q} \in \mathcal T^{\mathcal Q}$ such that $\sum_{z \in \mathcal N_\text{int}(\mathcal T_{q(t)})} \Delta \hat L_Y(z;\beta)>0$.
    
    Now consider $\mathcal S \neq \emptyset$. 
    By hypothesis, there exists a tree $\mathcal T_{ q^*} \in \mathcal T^{\mathcal Q}$ such  that
	\begin{equation*}
	\sum_{z \in \mathcal N_{\text{int}}(\mathcal T_{ q^*(t')})}\Delta \hat L_Y(z;\beta) = \hat Q_Y(t';\beta),  ~~ \forall t' \in \mathcal S.
	\end{equation*}
	Consider a  tree $\mathcal T_{ q} \in \mathcal T^{\mathcal Q}$ such that $\mathcal T_{ q(t)}$ has the properties
	$$\mathcal N_\text{int}(\mathcal T_{ q(t)}) = \left\{ t \right\} \bigcup\limits_{t' \in \mathcal S} \mathcal N_\text{int}(\mathcal T_{ q^*(t')}),$$
	and 
	$$\mathcal N_\text{leaf}(\mathcal T_{ q(t)}) = \left( \mathcal C (t) \setminus \mathcal S \right) \bigcup\limits_{t' \in \mathcal S}\mathcal N_\text{leaf}(\mathcal T_{ q^*(t')}).$$
	Therefore, using the fact that $\sum_{z \in \mathcal N_{\text{int}}(\mathcal T_{ q^*(t')})}\Delta \hat L_Y(z;\beta) = \hat Q_Y(t';\beta)$, for all $t' \in \mathcal S$, we have 
	\begin{align*}
	\sum_{z \in \mathcal N_\text{int}(\mathcal T_{ q(t)})}\Delta \hat L_Y(z;\beta) &= \Delta \hat L_Y(t;\beta) + \sum_{t' \in \mathcal S}\sum_{z \in \mathcal N_\text{int}(\mathcal T_{q^*(t')})}\Delta \hat L_Y(z;\beta), \\
	&= \Delta \hat L_Y(t;\beta) + \sum_{t' \in \mathcal S} \hat Q_Y(t';\beta).
	\end{align*}
	Also note that $\hat Q_Y(t';\beta) = 0$ for all $t' \in \mathcal C (t) \setminus \mathcal S$ and hence,
	\begin{align*}
	\sum_{z \in \mathcal N_\text{int}(\mathcal T_{ q(t)})}& \Delta \hat L_Y(z;\beta) = \Delta \hat L_Y(t;\beta) + \sum_{t' \in \mathcal S}\hat Q_Y(t';\beta) + \sum_{t' \in \left\{ \mathcal C(t) \setminus \mathcal S \right\}} \hat Q_Y(t';\beta).
	\end{align*}
	Furthermore, note that from Definition \ref{def:node_wise_Q}, if $\hat Q_Y(t;\beta)>0$ then
	\begin{equation*}
	    \hat Q_Y(t;\beta)=\Delta \hat L_Y(t;\beta) + \sum_{t' \in \mathcal C(t)}\hat Q_Y(t';\beta),
	\end{equation*}
	and thus,
	\begin{equation*}
	\sum_{z \in \mathcal N_\text{int}(\mathcal T_{ q(t)})} \Delta \hat L_Y(z;\beta) = \hat Q_Y(t;\beta) > 0.
	\end{equation*}
	
	Therefore, it follows that if $\hat Q_Y(t;\beta)>0$, there exists a tree such that  $\sum_{z \in \mathcal N_\text{int}(\mathcal T_{ q(t)})} \Delta \hat L_Y(z;\beta)>0$ and $ \sum_{z \in \mathcal N_\text{int}(\mathcal T_{ q(t)})} \Delta \hat L_Y(z;\beta)$ $= \hat Q_Y(t;\beta)$.
	
	Furthermore, consider any $\mathcal T_{\tilde q} \in \mathcal T^{\mathcal Q}$ such that $\mathcal T_{\tilde q(t)} \neq \mathcal T_{q(t)}$.
	Then
	\begin{align*}
	&\sum_{z \in \mathcal N_\text{int}(\mathcal T_{\tilde q(t)})}\Delta \hat L_Y(z;\beta) = \Delta \hat L_Y(t;\beta) + \sum_{t'\in \left\{\mathcal C(t) \cap \mathcal N_\text{int}(\mathcal T_{\tilde q(t)}) \right\}}\left( \sum_{z \in \mathcal N_\text{int}(\mathcal T_{\tilde q(t')})}\Delta \hat L_Y(z;\beta)\right).
	\end{align*}
	Note that $t' \in \mathcal N_k (\mathcal T_{\mathcal W})$ and that
	\begin{equation*}
	    \sum_{z \in \mathcal N_\text{int}(\mathcal T_{\tilde q(t')})}\Delta \hat L_Y(z;\beta) \le \hat Q_Y(t';\beta).
	\end{equation*}
	Consequently, 
	\begin{align*}
	\sum_{z \in \mathcal N_\text{int}(\mathcal T_{\tilde q(t)})}\Delta \hat L_Y(z;\beta) &\leq \Delta \hat L_Y(t;\beta) + \sum_{t'\in \left\{ \mathcal C(t) \cap \mathcal N_{\text{int}}(\mathcal T_{\tilde q(t)}) \right\}}\hat Q_Y(t';\beta), \\
	&\leq \Delta \hat L_Y(t;\beta) + \sum_{t'\in \mathcal C (t)}\hat Q_Y(t';\beta), \\
	&= \hat Q_Y(t;\beta).
	\end{align*}
	\\
	\noindent  $\left( \Leftarrow \right)$ Let $t \in \mathcal N_{k-1}(\mathcal T_{\mathcal W})$ and assume that there exists a tree $\mathcal T_{q} \in \mathcal{T}^{\mathcal Q}$ such that 
	\begin{equation*}
	\sum_{z \in \mathcal N_{\text{int}}(\mathcal T_{q(t)})}\Delta \hat L_Y(z;\beta) > 0,
	\end{equation*}
	and consider any $t' \in \mathcal N_\text{int}(\mathcal T_{q(t)}) \cap \mathcal C(t) \subset \mathcal N_{k}(\mathcal T_{\mathcal W})$.
	From the hypothesis  we have that
	\begin{equation*}
	\sum_{z \in \mathcal N_\text{int}(\mathcal T_{q(t')})}\Delta \hat L_Y(z;\beta) \le \hat Q_Y(t';\beta).
	\end{equation*}
	Therefore,
	\begin{align*}
	\sum_{z \in \mathcal N_\text{int}(\mathcal T_{q(t)})}\Delta \hat L_Y(z;\beta) = \Delta \hat L_Y(t;\beta) + \sum_{t'\in \left\{\mathcal C(t) \cap \mathcal N_\text{int}(\mathcal T_{q(t)}) \right\} }\left(\sum_{z \in \mathcal N_\text{int}(\mathcal T_{q(t')})}\Delta \hat L_Y(z;\beta)   \right), \\
	\end{align*}
	which yields
	\begin{align*}
	0< &\sum_{z \in \mathcal N_\text{int}(\mathcal T_{q(t)})}\Delta \hat L_Y(z;\beta) \le \Delta \hat L_Y(t;\beta) + \sum_{t' \in \left\{ \mathcal C(t) \cap \mathcal N_\text{int}(\mathcal T_{q(t)})\right\}} \hat Q_Y(t';\beta) +  \underbrace{\sum_{t' \in \left\{ \mathcal C(t) \setminus \mathcal N_\text{int}(\mathcal T_{q(t)}) \right\} } \hat Q_Y(t';\beta)}_\text{$\ge 0$}.
	\end{align*}
	Hence,
	\begin{equation*}
	0 < \Delta \hat L_Y(t;\beta) + \sum_{t' \in \mathcal C(t)}\hat Q_Y(t';\beta) \le \hat Q_Y(t;\beta).
	\end{equation*}
	Therefore, the existence of a tree $\mathcal T_{q} \in \mathcal{T}^{\mathcal Q}$ with $\sum_{z \in \mathcal N_\text{int}(\mathcal T_{q(t)})}\Delta \hat L_Y(z;\beta)>0$ where $t \in \mathcal N_{k-1} (\mathcal T_{\mathcal W})$ implies $\hat Q_Y(t;\beta)>0$. 
	
	\vspace{0.2cm}
	\noindent Thus, we have shown that the lemma holds for $k-1$ and for all $t \in \mathcal{N}_{k-1}(\mathcal{T}_{\mathcal{W}})$.

\subsection{Proof of Lemma \ref{lem:Qsearch2}}

	Let $t \in \mathcal N(\mathcal T_{\mathcal W})$ be any node such that $t \in \mathcal N_{\text{leaf}}(\mathcal T_{q'}) \cap \mathcal N_{\text{int}}(\mathcal T_{q^*})$,
	where $\mathcal T_{q'} \subset \mathcal T_{q^*}$.
	Note that $\hat Q_Y(n;\beta) > 0$ for all  $n \in \mathcal N_{\text{int}}(\mathcal T_{q^*(t)})$ and $\hat Q_Y(n;\beta) = 0$ for all $n \in \mathcal N_{\text{leaf}}(\mathcal T_{q^*(t)})$, which follows from the design of the Q-tree search algorithm.
	Thus, we have that
	\begin{equation*} 
	\sum_{z \in \mathcal N_{\text{int}}(\mathcal T_{q^*(t)})} \Delta \hat L_Y(z;\beta) = \hat Q_Y(t;\beta) > 0,
	\end{equation*}
	which holds for all $t \in \mathcal N_{\text{leaf}}(\mathcal T_{q'}) \cap \mathcal N_{\text{int}}(\mathcal T_{q^*})$.
	Furthermore, using \eqref{eq:lagrangianNodeDeltaL},
	\begin{align*}
	L_Y(\mathcal T_{q'};\beta) + \hspace*{-7mm} \sum_{t \in \left\{ \mathcal N_{\text{leaf}}(\mathcal T_{q'}) \cap \mathcal N_{\text{int}}(\mathcal T_{q^*}) \right\}} &\sum_{z \in \mathcal N_{\text{int}}(\mathcal T_{q^*(t)})} \Delta \hat L_Y(z;\beta) = L_Y(\mathcal T_{q^*};\beta).
	\end{align*}
	The above is equivalent to
	\begin{equation*}
	L_Y(\mathcal T_{q'};\beta) + \hspace*{-7mm}  \sum_{t \in \left\{ \mathcal N_{\text{leaf}}(\mathcal T_{q'}) \cap \mathcal N_{\text{int}}(\mathcal T_{q^*}) \right\} } \hat Q_Y(t;\beta) = L_Y(\mathcal T_{q^*};\beta).
	\end{equation*}
	Lastly, it is known that Q-tree search did not terminate at $\mathcal T_{q'}$. 
	Thus, $\sum_{t \in \left\{ \mathcal N_{\text{leaf}}(\mathcal T_{q'}) \cap \mathcal N_{\text{int}}(\mathcal T_{q^*})\right\} } \hat Q_Y(t;\beta) > 0$, where $N_{\text{leaf}}(\mathcal T_{q'}) \cap \mathcal N_{\text{int}}(\mathcal T_{q^*}) \neq \emptyset$ if  $\mathcal T_{q'} \ne \mathcal T_{q^*}$, and therefore
	\begin{equation*}
	L_Y(\mathcal T_{q'};\beta) < L_Y(\mathcal T_{q^*};\beta).
	\end{equation*}

\subsection{Proof of Theorem \ref{thm:Qsearch3}}
%
%
	Let $t \in \mathcal N_{\text{int}}(\mathcal T_{\tilde q})$ and consider the tree $\mathcal T_{\bar q} \in \mathcal T^{\mathcal Q}$ with node set $\mathcal N(\mathcal T_{\bar q}) =\{t\}\cup \mathcal N(\mathcal T_{\tilde q}) \setminus \mathcal N (\mathcal T_{\tilde q(t)})$.
	We have from \eqref{eq:lagrangianNodeDeltaL} that
	\begin{equation*}
	L_Y(\mathcal T_{\bar q};\beta) = \sum_{z \in \left\{ \mathcal{N}_{\text{int}}(\mathcal T_{\tilde q}) \setminus \mathcal{N}_{\text{int}}(\mathcal T_{\tilde q(t)}) \right\}  } \Delta \hat L_Y(z;\beta).
	\end{equation*}
	From the above expression and \eqref{eq:seqCharL} and \eqref{eq:deltaLnodeRelaton}, we have
	\begin{equation*}
	L_Y(\mathcal T_{\tilde q};\beta) = L_Y(\mathcal T_{\bar q};\beta) + \sum_{z\in {\mathcal N}_{\text{int}}(\mathcal T_{\tilde q(t)})} \Delta \hat L_Y(z;\beta).
	\end{equation*}
	Since $\mathcal T_{\tilde q}$ is minimal, for any subtree $\mathcal T_{\bar q}$ we have  $L_Y(\mathcal T_{\bar{q}};\beta) < L_Y(\mathcal T_{\tilde q};\beta)$, and therefore 
	\begin{equation*}
	    \sum_{z\in {\mathcal N}_{\text{int}}(\mathcal T_{\tilde q(t)})} \Delta \hat L_Y(z;\beta) > 0,~~~\forall t\in \mathcal N_\text{int}(\mathcal T_{\tilde q}).
	\end{equation*}
	Hence, from Lemma \ref{lem:qtree2}, $\hat Q_Y(t;\beta)  > 0$ for all $t \in \mathcal N_\text{int}(\mathcal T_{\tilde q})$.
	Thus, all nodes in $\mathcal N_\text{int}(\mathcal T_{\tilde q})$ are expanded in $\mathcal T_{q^*}$, which implies that $\mathcal T_{q^*} \supseteq \mathcal T_{\tilde q}$. 
	Then, either $\mathcal T_{q^*} = \mathcal T_{\tilde q}$, which implies $L_Y(\mathcal{T}_{\tilde q};\beta) = L_Y(\mathcal{T}_{ q^*};\beta)$, or $\mathcal T_{q^*} \supset \mathcal T_{\tilde q}$, which, from Lemma \ref{lem:Qsearch2}, implies that $L_Y(\mathcal{T}_{\tilde q};\beta) < L_Y(\mathcal{T}_{ q^*};\beta)$.
	However, since $\mathcal{T}_{\tilde q}$ is optimal, we have $L_Y(\mathcal{T}_{\tilde q};\beta) \ge L_Y(\mathcal{T}_{ q^*};\beta)$, leading to a contradiction. 
	Thus,  $\mathcal T_{q^*} = \mathcal T_{\tilde q}$ and  consequently $L_Y(\mathcal{T}_{\tilde q};\beta) = L_Y(\mathcal{T}_{ q^*};\beta)$.
    
\subsection{Proof of Proposition \ref{prop:limitDeltaL}}
	Assume $\beta > 0$, $t \in \mathcal N_{\text{int}}(\mathcal T_{\mathcal W})$ and $p(x) = {\varepsilon}/{N}$ for all $x\in \mathcal N_{\text{leaf}}(\mathcal T_{\mathcal W(t)})$ with $N = \lvert \mathcal N_{\text{leaf}}(\mathcal T_{\mathcal W(t)}) \rvert$.
	By \eqref{eq:delta_L_agglomerative_IB} and Definition \ref{def:deltaLhat}, we have 
	\begin{equation*}
	\Delta \hat L(t;\beta )  = \p(t)\left[ \js(\p(y|t'_{1}),\ldots,\p(y|t'_{\lvert \mathcal C(t) \rvert })) - \frac{1}{\beta}\H(\Pi) \right],
	\end{equation*}
	where, without loss of generality, $\left\{t'_1, \ldots, t'_{\lvert \mathcal C(t) \rvert } \right\} = \mathcal C(t)$.
	Moreover, since $p(t|x)$ is deterministic,  
	\begin{equation*}
	p(t) = \sum_{x\in \mathcal N_{\text{leaf}}(\mathcal T_{\mathcal W})} p(t|x)p(x) = \sum_{x\in \mathcal N_{\text{leaf}}(\mathcal T_{\mathcal W(t)})} p(x) = \varepsilon,
	\end{equation*}
	and since $p(x) = {\varepsilon}/{N}$ for all $x \in \mathcal N_{\text{leaf}}(\mathcal T_{\mathcal W(t)})$, it follows that
	\begin{equation*} \label{eq:propProofpt'}
	p(t') = \frac{\varepsilon}{\lvert \mathcal C(t) \rvert}, \quad  t' \in \mathcal C(t).
	\end{equation*}
	Consequently, 
	\begin{equation*}
	\Pi = \left\{\frac{\p(t'_{1})}{\p(t)},\ldots,\frac{\p(t'_{\lvert \mathcal C(t) \rvert})}{\p(t)} \right\} = \left\{ \frac{1}{\lvert \mathcal C(t) \rvert},\ldots, \frac{1}{\lvert \mathcal C(t) \rvert}\right\},
	\end{equation*}
	and therefore,
	\begin{equation} \label{eq:propProofEntrEps}
	H(\Pi) = \log \lvert \mathcal C(t) \rvert.
	\end{equation}
	Now define
	\begin{equation*}
	a_{t'}(y) \triangleq \sum_{x \in \mathcal N_{\text{leaf}}\left( \mathcal T_{\mathcal W(t')}\right)} p(x,y),
	\end{equation*}
	and 
	\begin{equation*}
	a_{t}(y) \triangleq \sum_{x \in \mathcal N_{\text{leaf}}\left( \mathcal T_{\mathcal W(t)}\right)} p(x,y),
	\end{equation*}
	where $y \in \Omega_Y$ and $t' \in \mathcal C(t)$.
	Thus, from the definition of $a_{t'}(y)$ and $a_t(y)$,
	\begin{equation} \label{eq:propProofatprimeSum}
	\sum_y a_{t'}(y) = \frac{\varepsilon}{\lvert \mathcal C(t) \rvert},
	\end{equation}
	and
	\begin{equation*}
	\sum_y a_{t}(y) = \varepsilon,
	\end{equation*}
	for all $t' \in \mathcal C(t)$.
	Since $\mathcal N_{\text{leaf}}(\mathcal T_{\mathcal W(t')}) \subseteq \mathcal N_{\text{leaf}}(\mathcal T_{\mathcal W(t)})$, it follows that $0\leq a_{t'}(y) \leq a_t(y) \leq \varepsilon$.
	Thus, for $t' \in \mathcal C(t)$ we have, from the definition of the KL-divergence, 
	\begin{equation*}
	\dkl(p(y|t'),p(y|t)) = \sum_y p(y|t')\log\frac{p(y|t')}{p(y|t)},
	\end{equation*}
	where 
	\begin{equation*}
	p(y|t) = \frac{1}{p(t)}\sum_{x \in \mathcal N_{\text{leaf}(\mathcal T_{\mathcal W(t)})}}p(x,y) = \frac{1}{\varepsilon} a_{t}(y),
	\end{equation*}
	and similarly,
	\begin{equation*}
	p(y|t') = \frac{1}{p(t')}\sum_{x \in \mathcal N_{\text{leaf}(\mathcal T_{\mathcal W(t')})}}p(x,y) = \frac{\lvert \mathcal C(t) \rvert}{\varepsilon} a_{t'}(y).
	\end{equation*}
	Hence,
	\begin{align}
	\dkl(p(y|t&'),p(y|t)) = \sum_y p(y|t') \log \frac{\lvert \mathcal C(t) \rvert a_{t'}(y)}{a_t(y)},\nonumber \\
	&= \log \lvert \mathcal C(t) \rvert + \sum_y p(y|t')\log \frac{a_{t'}(y)}{a_t(y)}, \nonumber\\
	&= \log \lvert \mathcal C(t) \rvert + \frac{\lvert \mathcal C(t) \rvert}{\varepsilon} \sum_y a_{t'}(y)\log \frac{a_{t'}(y)}{a_t(y)} \label{eq:propProofKLdiv1}.
	\end{align}
	Since $ 0\leq a_{t'}(y) \leq a_t(y)$ for all $y \in \Omega_Y$ we have from \eqref{eq:propProofatprimeSum} and \eqref{eq:propProofKLdiv1} that
	\begin{align*}
	\frac{\lvert \mathcal C(t) \rvert}{\varepsilon} \sum_y a_{t'}(y)\log \frac{a_{t'}(y)}{a_t(y)} &\leq \frac{\lvert \mathcal C(t) \rvert}{\varepsilon} \sum_y a_{t'}(y)\log \frac{a_t(y)}{a_t(y)}, \\
	&= \frac{\lvert \mathcal C(t) \rvert}{\varepsilon} \log\left(1\right) \sum_y a_{t'}(y), \\
	&= 0.
	\end{align*}
	Thus, from the previous expression, along with \eqref{eq:propProofKLdiv1}, it follows that
	\begin{equation} \label{eq:propProofKLbound}
	0 \leq \dkl(p(y|t'),p(y|t)) \leq  \log \lvert \mathcal C(t) \rvert , ~~~\forall t' \in \mathcal C(t).
	\end{equation}
	Using \eqref{eq:propProofKLbound} and the definition of JS-divergence, we see that 
	\begin{align*}
	\js(\p(y|t'_{1}),\ldots,\p(y|t'_{\lvert \mathcal C(t) \rvert }) &= \sum_{i=1}^{\lvert \mathcal C(t) \rvert} \Pi(i)\dkl(p(y|t'_i),p(y|t)),\\
	&\leq \log \lvert \mathcal C(t) \rvert.
	\end{align*}
	Therefore, from the non-negativity of the JS-divergence as well as \eqref{eq:propProofEntrEps} and \eqref{eq:propProofKLbound} we have,
	\begin{align*}
	- \frac{1}{\beta}\varepsilon \log\lvert \mathcal C(t) \rvert \leq \p(t)\big[ \js(&\p(y|t'_{1}),\ldots,\p(y|t'_{\lvert \mathcal C(t) \rvert })) - \frac{1}{\beta}\H(\Pi) \big] \leq \frac{\beta - 1}{\beta} \varepsilon \log \lvert \mathcal C(t) \rvert .
	\end{align*}
	Now taking the limit as $\varepsilon \to 0^{+}$ yields $\lim_{\varepsilon \to 0^+} \Delta \hat L(t;\beta) = 0$ for all $\beta > 0$.


\end{document}